\documentclass[10pt,twocolumn,letterpaper]{article}

\usepackage{cvpr}
\usepackage{times}
\usepackage{epsfig}
\usepackage{amsmath}
\usepackage{amssymb}

\usepackage{multirow}
\usepackage{array, booktabs}
\usepackage{arydshln}
\usepackage{calrsfs}
\usepackage[cal=boondoxo]{mathalfa}
\usepackage[mathscr]{euscript}

\usepackage{makecell}

\usepackage{authblk}

\usepackage{kotex}
\usepackage{color}
\DeclareMathAlphabet{\pazocal}{OMS}{zplm}{m}{n}

\usepackage{array,multirow}

\usepackage[ruled,vlined, linesnumbered]{algorithm2e} 

\usepackage{subcaption}
\usepackage{comment}

\definecolor{greyblue}{rgb}{0.1,0.6,0.5}

\definecolor{pp}{rgb}{0.6,0.0,0.6}

\definecolor{ban}{rgb}{0.08,0.70,0.75}

\definecolor{tan}{rgb}{0.578,0.450,0.015}

\definecolor{dodgerblue}{rgb}{0.12, 0.56, 1.0}

\newcommand\RED[1]{\textcolor{red}{#1}}
\newcommand\BLUE[1]{\textcolor{blue}{#1}}

\newcolumntype{L}[1]{>{\raggedright\arraybackslash}p{#1}}
\newcolumntype{C}[1]{>{\centering\arraybackslash}p{#1}}
\newcolumntype{R}[1]{>{\raggedleft\arraybackslash}p{#1}}

\definecolor{Gray}{gray}{0.85}
\newcolumntype{h}{>{\columncolor{Gray}}c}

\usepackage{hyperref}
\hypersetup{pagebackref=true,breaklinks=true,letterpaper=true,colorlinks,bookmarks=false}

\cvprfinalcopy 

\def\cvprPaperID{18} 

\ifcvprfinal\fi
\begin{document}

\title{CLEval: Character-Level Evaluation for Text Detection and Recognition Tasks}


\author[1]{Youngmin Baek}
\author[1]{Daehyun Nam}
\author[1]{Sungrae Park}
\author[1]{Junyeop Lee}
\author[1]{\protect\\Seung Shin}
\author[1]{Jeonghun Baek}
\author[2]{Chae Young Lee}
\author[1]{Hwalsuk Lee\thanks{Corresponding author.}}
\affil[1]{Clova AI Research, NAVER Corp.} \affil[2]{Yale University}

\renewcommand\Authands{ and }

\maketitle

\begin{abstract}
Despite the recent success of text detection and recognition methods, existing evaluation metrics fail to provide a fair and reliable comparison among those methods. In addition, there exists no end-to-end evaluation metric that takes characteristics of OCR tasks into account. Previous end-to-end metric contains cascaded errors from the binary scoring process applied in both detection and recognition tasks. Ignoring partially correct results raises a gap between quantitative and qualitative analysis, and prevents fine-grained assessment. Based on the fact that character is a key element of text, we hereby propose a Character-Level Evaluation metric (CLEval). In CLEval, the \textit{instance matching} process handles split and merge detection cases, and the \textit{scoring process} conducts character-level evaluation. By aggregating character-level scores, the CLEval metric provides a fine-grained evaluation of end-to-end results composed of the detection and recognition as well as individual evaluations for each module from the end-performance perspective. We believe that our metrics can play a key role in developing and analyzing state-of-the-art text detection and recognition methods. The evaluation code is publicly available at \url{https://github.com/clovaai/CLEval}.

\end{abstract}

\vspace{-0.1cm}

\section{Introduction}





\begin{figure}[t!]
    \centering
    \begin{subfigure}{\linewidth}
      \centering
      \includegraphics[width=\linewidth]{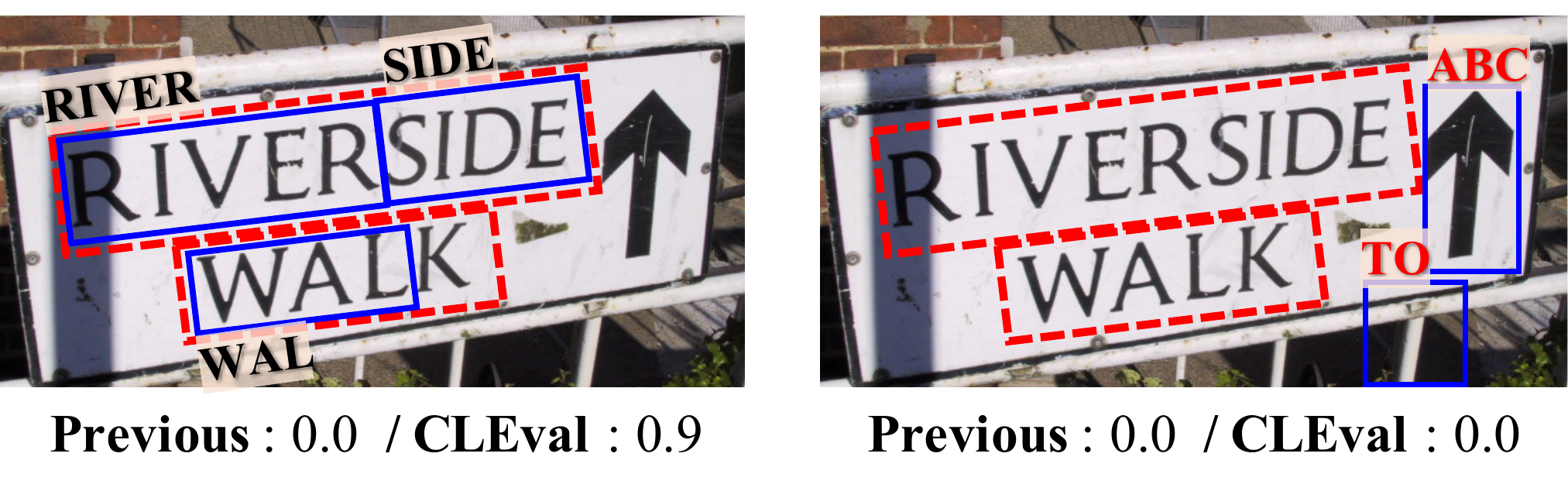}
      \vspace{-2em}
      \caption{Scores on problematic \emph{detection} examples}
      \vspace{0.75em}
      \label{fig:intro_a}
    \end{subfigure}
    \begin{subfigure}{\linewidth}
      \centering
      \includegraphics[width=\linewidth]{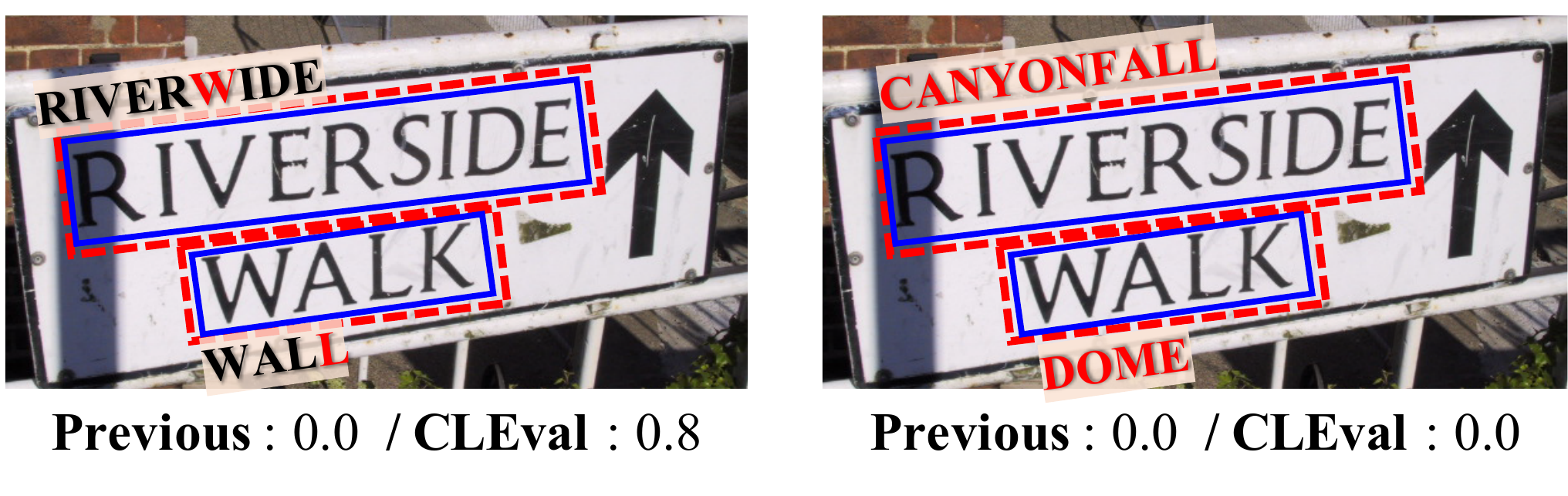}
      \vspace{-2em}
      \caption{Scores on problematic \emph{recognition} examples}
      \label{fig:intro_b}
    \end{subfigure}
    \caption{Comparison of CLEval with previous end-to-end metric\cite{karatzas2015icdar}. Figures in the left column are slightly wrong cases while figures in the right are completely wrong cases. \RED{Red}: GT. \BLUE{Blue}: detection. Texts indicate the recognized results of the detection box, and texts in \RED{red} are incorrect.}
    \label{fig:intro}
    \vspace{-1.5em}
\end{figure}

Along with the progress in the field of machine learning, the performances of text detectors and recognizers have remarkably improved over the past few years~\cite{tian2016ctpn, zhou2017east, shi2017seglink, baek2019craft, baek2019wrong, long2018scene}. However, existing detection and recognition evaluation metrics fail to provide a fair and reliable comparison among those methods. Especially when evaluating end-to-end models, errors are aggregated due to unconvincing measurements in each part. Fig. \ref{fig:intro} illustrates common problems encountered in end-to-end detection and recognition modules. The previously used instance-level binary scoring process assigns a value of 0 on both decent(left figures) and wrong(right figures) results.



%
%

\begin{figure*}[t!]
 \begin{subfigure}{.25\linewidth}
 \centering
 \includegraphics*[width=0.9\linewidth, clip=true]{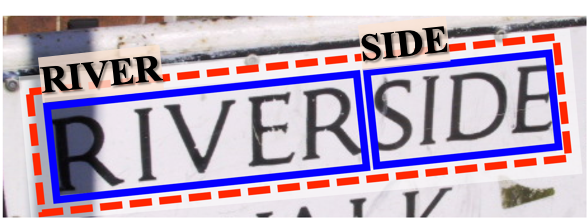}
 \caption{Split case}
 \end{subfigure}%
 \begin{subfigure}{.25\linewidth}
 \centering
 \includegraphics*[width=0.9\linewidth, clip=true]{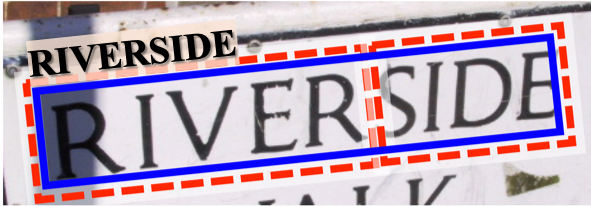}
 \caption{Merged case}
 \end{subfigure}%
 \begin{subfigure}{.25\linewidth}
 \centering
 \includegraphics*[width=0.9\linewidth, clip=true]{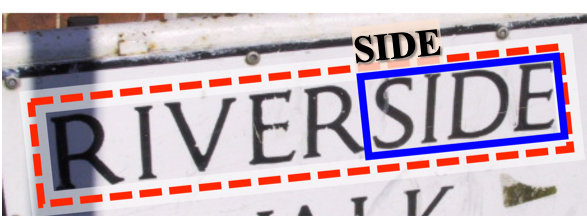}
 \caption{Missing characters}
 \end{subfigure}%
 \begin{subfigure}{.25\linewidth}
 \centering
 \includegraphics*[width=0.9\linewidth, clip=true]{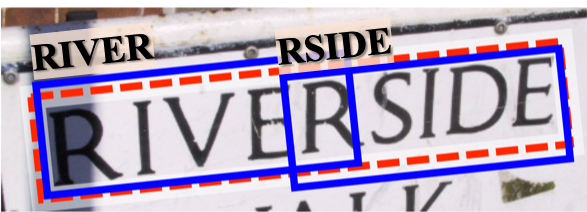}
 \caption{Overlapping characters}
 \end{subfigure}%
 \vspace{-3mm}
 \caption{Examples of issues for the fair evaluation of text detection and recognition tasks. (a) and (b) correspond to the issue of granularity, (c) and (d) are related to the issue of correctness.}
  \label{fig:issues}
  \vspace{-3mm}
\end{figure*}


To better understand where established text detection and recognition evaluation metrics fail, a closer look into the intrinsic nature of texts is required. At a fundamental level, text consists of words, which can further be decomposed into an array of characters. This character array embodies two intrinsic characteristics: its sequential nature and its content. Text detection's goal of locating words can then be reinterpreted as finding the area that encapsulates the right sequence and content in a group of characters. The degree to which the right sequence and content are recognized within the detection area denotes the \textit{granularity} and \textit{correctness} issues, respectively. A more specific explanation of these attributes follows -- with examples of what they measure in Figure~\ref{fig:issues}.

\textit{Granularity} is the degree to which the text detection model captures the sequence of characters in exactly one word as one unbroken sequence. Split detection results (Figure~\ref{fig:issues}(a)) break the character sequence in the word and merged detection results (Figure~\ref{fig:issues}(b)) fail to capture exactly one word. Both cases incur penalties proportionate to the number of splits or merges per word.

\textit{Correctness} is the degree to which the text detection and recognition model captures the content of word. Specifically each character in that word must be detected and recognized exactly once with its right order. A penalty is incurred proportionate to the number of missing or overlapping characters in both detection and recognition results (Figure~\ref{fig:issues}(c, d)). 

Majority of the public datasets provide a word-level annotation since each word contains a semantic meaning. 
However, as Fig. \ref{fig:intro} shows, evaluating word-level boxes with a predefined threshold provokes various issues. Despite having appropriate box predictions, the binary scoring process discards acceptable prediction results and produces unexplainable scores. To provide more detailed interpretation of the models, recent studies have adopted a character-level evaluation process \cite{lee2019tedeval, lee2019popeval}. Inspired by them, our proposed metric, named as CLEval (Character-Level Evaluation), is designed to perform end-to-end evaluations without explicit character annotations. The method adopts two key components; instance matching process and character scoring process. The instance matching process solves granularity issues by pairing all possible GT and detection boxes that share at least one character, and the character-level scoring process solves correctness issues by calculating the longest common subsequence between GT and predicted transcriptions.



The CLEval metric is primarily designed to evaluate end-to-end tasks, but it can also be applied to individual detection and recognition modules. Interpretation of each module is valuable since it allows us to discover how each component affects overall performance. Assuming that the characters are evenly placed within a word box, the detection evaluation is conducted using pseudo-character center positions. This method was first proposed by \cite{lee2019tedeval}, but we further developed the idea to handle end-to-end models based on the same scoring policy. The proposed metric also provides quantitative indicators of recognition modules by measuring correctly recognized words within detection boxes.

The main contributions of this paper can be summarized as follows. 1) We propose a character-level evaluation metric that is favorable qualitative view without character-level annotations.
2) We define granularity and correctness issues, and solve them by performing instance matching and character scoring processes. 3) We propose a unified protocol that could evaluate not only end-to-end tasks, but also individual detection and recognition modules.

\begin{figure*}[t!]
 \centering
 \includegraphics*[width=0.9\linewidth, clip=true]{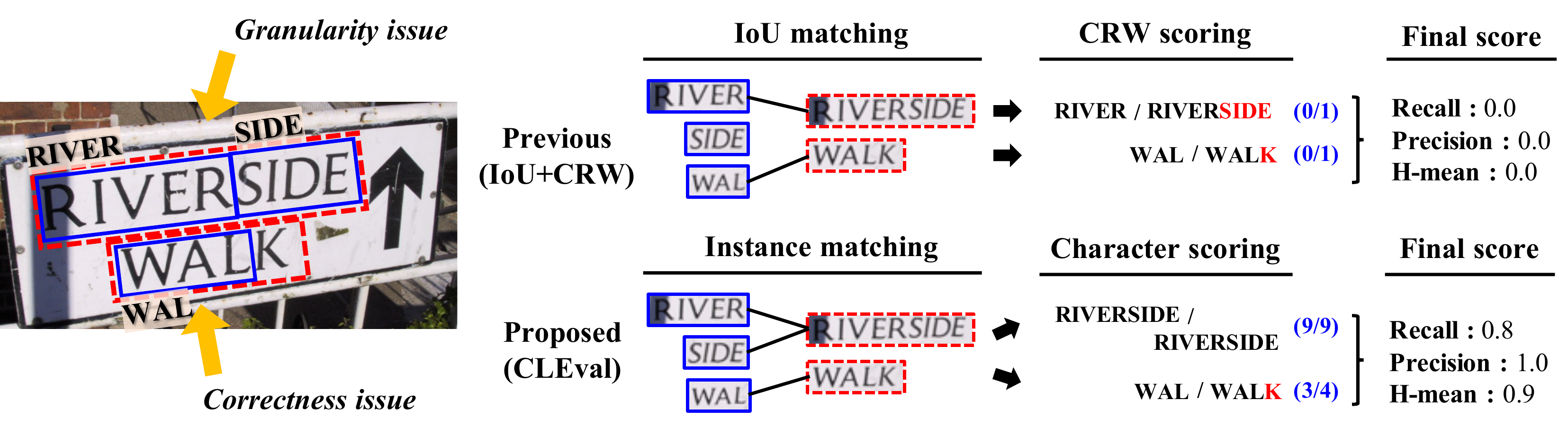} 
 \caption{Overall evaluation process of IoU+CRW and CLEval metrics. In IoU+CRW, only two bounding boxes are matched through 0.5 IoU threshold, and partially detected texts are discarded in CRW scoring process. In CLEval, all detection boxes are matched, and correct numbers of characters are reflected to the final evaluation.}
  \label{fig:overview}
\end{figure*}

\section{Related works}


\subsection{Detection evaluation} \label{detection}

\vspace{1mm}
\noindent {\bf Intersection-over-Union (IoU) }
IoU metric originally comes from object detection task such as Pascal VOC \cite{everingham2015pascal}. IoU accepts detections that match the ground truth (GT) box in an exclusive one-to-one manner only when the overlapping region satisfy the predefined threshold. Although IoU is the most widely used evaluation metric thanks to its simplicity, its behavior is clearly not suitable for evaluating texts as argued by  \cite{calarasanu2016good, nguyenstate}. IoU cannot handle granularity and correctness issues, which is critical for OCR tasks. 

\vspace{1mm}
\noindent {\bf DetEval }
DetEval \cite{wolf2013deteval} was designed to solve the granularity issue by allowing multiple relationships of a single bounding box.
Their matching processes are conducted by accepting one-to-one, one-to-many, and many-to-one relationships. However, each instance is evaluated based on both area recall and area precision thresholds. Area-based threshold not only causes correctness issues, but also has a limitation when trying to apply end-to-end evaluation.

\vspace{1mm}
\noindent {\bf Tightness-aware IoU (TIoU) }
Liu et al. recently suggested the TIoU metric that penalizes based on the occupation ratio between detection and ground truth. By doing that, TIoU tried to give the high score to more similar detection compared to the box of ground truth. 
The major weakness is that TIoU penalizes slight differences between the ground truth and detection, even if the recognition results of those detection boxes are same. This is far from the end user perspective, and it is unfair if the correct box got a different score under the TIoU metric due to the small perturbation of box size.

\vspace{1mm}
\noindent {\bf TedEval } A character-level evaluation metric for text detection has been proposed by \cite{lee2019tedeval}. The metric alleviates qualitative disagreements, but can only be used to evaluate text detection modules. We adopt the idea of using pseudo-characters, and apply a character-level evaluation process to evaluate end-to-end results.

\subsection{Recognition evaluation} \label{recognition}

\vspace{1mm}
\noindent {\bf Correctly Recognized Words (CRW)}
As its term suggests, the CRW is a binary score metric by judging correct answers when the transcription and the recognition result are exactly same. This method has a fundamental limit of a binary score system that fails to give different scores to an absurd recognition result and an almost accurate result.

\vspace{1mm}
\noindent {\bf Edit Distance~\cite{levenshtein1966binary}}
The Edit Distance (ED) method is a common algorithm used to quantify how dissimilar two given strings are. The ED of two strings is the minimum operation required to transform one string into another. In the most standard Levenshtein distance calculation, such an operation involves insertion, deletion, and substitution. 
Utilizing ED to evaluate scene text recognition model is considered reasonable in that the score reflects how well the model is performing with distance measures.
Longest Common Subsequence (LCS) is literally the longest common subsequence in a set of sequences, and is a specialized case of ED which only uses insertion and deletion operations~\cite{LCS-ED-Relation}. 


\subsection{End-to-end evaluation} \label{e2e}
\vspace{1mm}
\noindent {\bf IoU and CRW}
The IoU and CRW are strictly a cascaded evaluation metric. The detection stage filters out detection results whose IoU with the corresponding GT is below the threshold. Matches with IoU are judged by the CRW. Both metrics in each of the stages are reported to have hindrances for fine-grained assessment due to the binary chain scoring.

\vspace{1mm}
\noindent {\bf PopEval }
PopEval was proposed to make full use of text information from recognition results. Their character elimination process is simple yet good enough from a practical point of view. However, PopEval does not provide detection evaluation. We adopted the idea of character elimination, and enhanced by using a substring elimination scheme to mitigate the problem of ignoring the order of texts.



    

\section{Methodology}

Fig. \ref{fig:overview} shows comparison of our method with the IoU + CRW metric. Detection boxes in the word “RIVERSIDE” show the granularity issue, and boxes in the word “WALK” show the correctness issue. While previous metrics fail to accept decent prediction results, our metric successfully quantifies various conditions through the matching process and the scoring process. The matching process identifies instance-level pairs between GT and detected boxes, and the scoring process provides a final score by analyzing extracted statistics. 

\subsection{Matching process}
In this section, the matching process is explained in detail. First, to overcome the absence of character annotations, we adopt the idea from \cite{lee2019tedeval} and calculate Pseudo-Character Center(PCC) positions. GT and detection boxes are considered a match if they satisfy two conditions. Their overlapping regions between GT and detection must share at least one PCC in common and should also cover an adequate GT area. Any candidates that do not satisfy the conditions are filtered out from the matching process.

\subsubsection{Pseudo-Character Center (PCC)}

We first need to know the location of the characters to identify whether a character region is covered by a detection box. However, most of the public datasets only provide word-level bounding box annotations. To handle this issue, we synthetically generate Pseudo-Character Center(PCC) points using GT word box and transcription.
\begin{figure}[t]
    \centering
    \includegraphics[width=.8\linewidth]{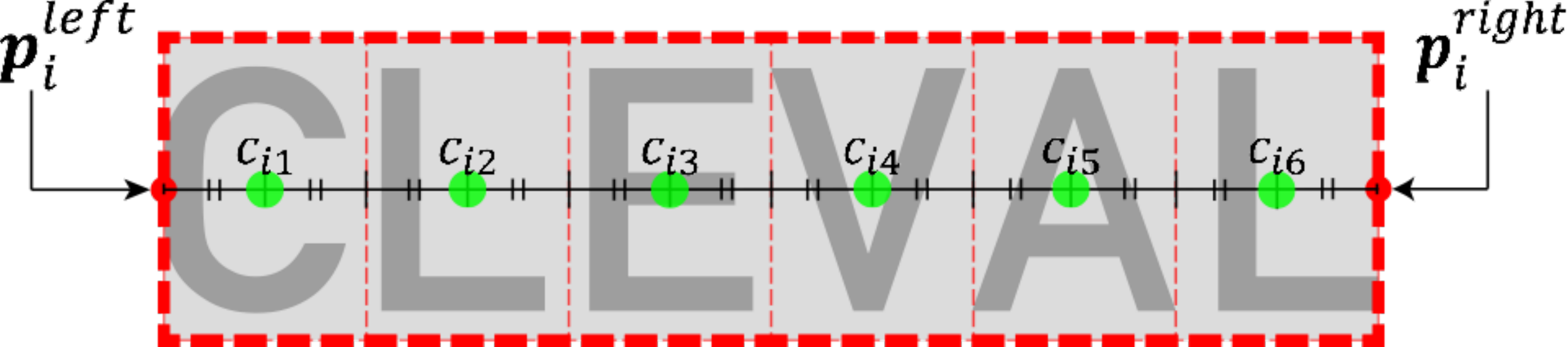}
    \caption{An example of computing PCC of $\text{G}_i$. Green dot: PCC. Red dash: pseudo character box. Grey: $\text{G}_i$.}
    \label{fig:pcc_generation}
\end{figure}
 Let $\boldsymbol{G}=\{G_1, ..., G_I\}$ be a set of GT boxes and $\boldsymbol{D}=\{D_1, ..., D_J\}$ be a set of detected boxes where $I$ and $J$ denote the size of each set. Each GT box, $\text{G}_i$, contains a word with $l^{G}_i$ characters. As shown in Fig. ~\ref{fig:pcc_generation}, we compute the $k$-th PCC of $\text{G}_i$ to obtain the positional information of the characters, 
\begin{equation}
	\label{eq:pcc}
    c^{k}_{i} = \left(\frac{2k-1}{2l_i^G}\right)p^{\text{left}}_i + \left(1-\frac{2k-1}{2l^G_i}\right) p^{\text{right}}_i
\end{equation}
where $p^{\text{left}}_i$ and $p^{\text{right}}_i$ indicate midpoints located on the left and right edges of $\text{G}_i$. The equation allows us to construct PCC points using both quadrilateral and polygon boxes. PCC point generation for polygon boxes is described in the Appendix.

\begin{figure}[t!]
    \centering
    \begin{subfigure}{.3\linewidth}
    \centering
    \includegraphics*[width=0.97\linewidth, clip=true]{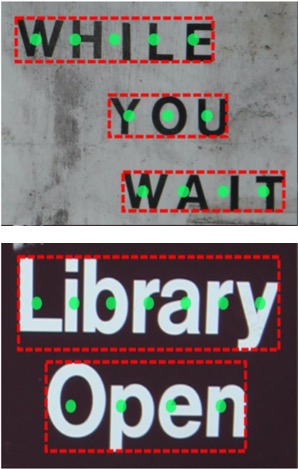}
    \caption{ICDAR2013\cite{karatzas2013icdar}}
    \end{subfigure}%
    \begin{subfigure}{.3\linewidth}
    \centering
    \includegraphics*[width=0.97\linewidth, clip=true]{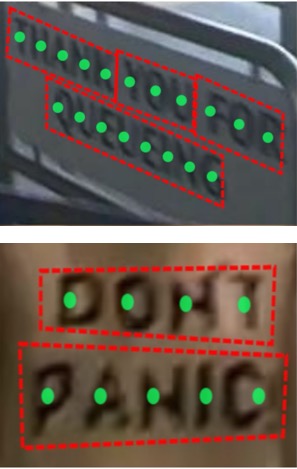}
    \caption{ICDAR2015\cite{karatzas2015icdar}}
    \end{subfigure}%
    \begin{subfigure}{.3\linewidth}
    \centering
    \includegraphics*[width=0.97\linewidth, clip=true]{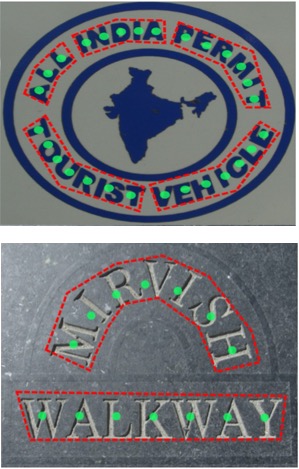}
    \caption{TotalText\cite{ch2017total}}
    \end{subfigure}%
 
    \caption{Visualization of PCC points on different ground-truth annotations.}
    \label{fig:pcc_examples}
    \vspace{-1.5em}
\end{figure}

The PCC points are generated under the assumption that the characters are evenly divided within a word box. However, constructed points may not be perfectly aligned with actual character positions because characters of different sizes coexist in the word image. Even with this ambiguity, our assumption works fairly well in most cases. Fig. ~\ref{fig:pcc_examples} shows constructed PCC points on real datasets.


\subsubsection{Matching based on character inclusion}

The first criterion for finding a match between GT and detection box is the inclusion of at least one PCC point. Here, we define character inclusion candidate, $\hat{m}^{k}_{ij}$, between $\text{D}_j$ and $\text{G}_i$ as
\begin{equation}
    \hat{m}^{k}_{ij} = \mathbb{I}( c^{k}_{i} \; \text{in} \; \text{D}_j)
\end{equation}
where $\mathbb{I}(\text{A})$ is a conditional function that gives a value of 1 when \text{A} is satisfied and 0 otherwise. If $\hat{m}^{k}_{ij}=1$, it is probable that $G_i$ and $D_j$ is matched since they share at least one PCC in common.


The second matching criterion is the area of intersection between detection box and GT text region. Since PCC is a single coordinate in the image, inclusion of a point does not guarantee good localization of the ground truth text region. In order to alleviate this ambiguity, detection boxes that cover small GT box regions are filtered out. To this end, the area precision of $\text{D}_j$ is defined as follow
\begin{equation}
	\label{eq:pcc}
    {AreaPrecision}_j = \frac{Area(\cup_{i \in \{i\|\exists k,\; \hat{m}^{k}_{ij}=1\}} ( \text{D}_j \cap \text{G}_i ) ) )}{Area(\text{D}_j)},
\end{equation}
where the union condition, $\{i\|\exists k,\; \hat{m}^{k}_{ij}=1\}$, indicates a set of GT boxes that contains at least one of its PCC points matched with the $\text{D}_j$. The matching process filters out candidates whose non-text region is larger than the text region. Therefore, the final box matching flags $M_{ij}$ are defined by considering the character inclusion flag $m^{k}_{ij}$ and its area precision as
\begin{equation}
\begin{split}
    m^{k}_{ij} & = \hat{m}^{k}_{ij}\times\mathbb{I}({AreaPrecision}_j > 0.5 ),\\
    M_{ij} & = \mathbb{I}(\sum_k m^{k}_{ij} > 0 ).
\end{split}
\end{equation}
To solve the granularity issue, we need to consider one-to-many and many-to-one cases. In our matching process, $AreaPrecision_j$ explicitly handles one-to-one and many-to-one cases by calculating the union of intersections between matched $G_i$s and $D_j$. In our metric, one-to-many match does not need to be processed since each split detection box is matched with one GT box by checking the inclusion of at least one character.





\subsubsection{Summarized matching statistics}

Table. \ref{tab:match} shows matching flags and statistics. The upper and lower cases mean box-level and the character-level instances, respectively. The subscript indicates the box index, and the superscript represents the character index. The statistics are written in the script font, and they represent row wise and column wise summations.
\begin{table}[h]
\vspace{-3mm}
    \fontsize{8}{8}\selectfont
    \tabcolsep=0.12cm
    \renewcommand*{\arraystretch}{1.2}
    \centering
    \begin{tabular}{c|c||c|c|c|c|c||c|c|c}
    \hline
    \multicolumn{2}{c||}{} & \multicolumn{2}{c|}{ $\text{D}_1$ } & ... & \multicolumn{2}{c||}{ $\text{D}_j$ } & \multicolumn{2}{c|}{Stat.} & Recall \\
    \hline \hline
    \multirow{4}{*}{ $\text{G}_1$ } & $c^{1}_{1}$ & \multirow{4}{*}{ $M_{11}$ } & $m^{1}_{11}$ & \multirow{4}{*}{ } & \multirow{4}{*}{ $M_{1j}$ } & $m^{1}_{1j}$ & \multirow{4}{*}{$\mathcal{G}_{1}$ } & $\mathcal{g}^{1}_{1}$ & \multirow{4}{*}{$R_1$ } \\
    & $c^{2}_{1}$ & & $m^{2}_{11}$ & & & $m^{2}_{1j}$ & & $\mathcal{g}^{2}_{1}$ & \\
    & ... & & ... & & & ... & & ... &  \\
    & $c^{k}_{1}$ & & $m^{k}_{11}$ & & & $m^{k}_{1j}$ & & $\mathcal{g}^{k}_{1}$ & \\
    \hline
    ... & ... & ... & ... & & ... & ... & ... & ... & ...\\
    \hline
    \multirow{4}{*}{ $\text{G}_i$ } & $c^{1}_{i}$ & \multirow{4}{*}{ $M_{i1}$ } & $m^{1}_{i1}$ & \multirow{4}{*}{ } & \multirow{4}{*}{ $M_{ij}$ } & $m^{1}_{ij}$ & \multirow{4}{*}{$\mathcal{G}_{i}$ } & $\mathcal{g}^{1}_{i}$ & \multirow{4}{*}{$R_i$ } \\
    & $c^{2}_{i}$ & & $m^{2}_{i1}$ & & & $m^{2}_{ij}$ & & $\mathcal{g}^{2}_{i}$ & \\
    & ... & & ... & & & ... & & ... &  \\
    & $c^{k}_{1}$ & & $m^{k}_{i1}$ & & & $m^{k}_{ij}$ & & $\mathcal{g}^{k}_{i}$ & \\
    \hline \hline
    \multicolumn{2}{c||}{Stat.} & $\mathcal{D}_{1}$ & $\mathcal{d}_{1}$ & & $\mathcal{D}_{j}$ & $\mathcal{d}_{j}$ & \multicolumn{3}{c}{}\\
    \cline{1-7}
    \multicolumn{2}{c||}{ Precision } & \multicolumn{2}{c|}{$P_1$ } &... & \multicolumn{2}{c||}{$P_j$ } & \multicolumn{3}{c}{} \\
    \hline
    \end{tabular}
    \vspace{-0.5em}
    \caption{Table of box-level and character-level matching flags and their statistics.}
    \label{tab:match}
\end{table}

Matching statistics are summarized in Table~\ref{tab:statistics}. The values are obtained using the character inclusion flag $m^{k}_{ij}$ and the box matching flag  $M_{ij}$. By aggregating matching statistics, we could identify total number of box matches and character inclusions. The value of $\mathcal{G}_i$ and $\mathcal{D}_j$ show the number of matched box candidates on each $\text{G}_i$ and $\text{D}_j$. The number $\mathcal{g}^{k}_{i}$ denotes matched detection boxes on $k$-th PCC point of $G_i$, and $\mathcal{d}_{j}$ shows the number of PCC points covered by a detection box $D_j$. These statistics are finally used to calculate character scores and granularity penalties.
\begin{table}[h]
    \tabcolsep=0.08cm 
    \renewcommand{\arraystretch}{1.5} 
    \fontsize{9}{8.5}\selectfont
    \centering
    \begin{tabular}{c|c|l}
        \hline
         Stat. & Eq. & Description\\
         \hline \hline \vspace{0.1em}
         $\mathcal{G}_i$ & $\sum\limits_j M_{ij}$ & number of $\boldsymbol{D}$ matched with $\text{G}_i$ \\
         \hline
         $\mathcal{D}_j$ & $\sum\limits_i M_{ij}$ & number of $\boldsymbol{G}$ matched with $\text{D}_j$ \\
         \hline
         $\mathcal{g}^{k}_{i}$ & $\sum\limits_j m^{k}_{ij}$ & number of $\boldsymbol{D}$ including $k$-th character of $\text{G}_i$ \\
         \hline
         $\mathcal{d}_{j}$ & $\sum\limits_{ik} m^{k}_{ij}$ & number of characters in $\text{G}_i$s matched with $\text{D}_j$ \\
         \hline
    \end{tabular}
    \vspace{-0.5em}
    \caption{Description of matching statistics.}
    \label{tab:statistics}
    \vspace{-1.5em}
\end{table}
\subsection{Scoring process} \label{scoring process}
Once the matching candidates are obtained, we now evaluate character-level correctness. Eq. \ref{eq:score_policy} is the ground rule to calculate recall and precision.
\begin{equation} \label{eq:score_policy}
\begin{split}
\textit{Score} = \frac{\textit{CorrectNum} - \textit{GranulPenalty}}{\textit{TotalNum}}
\end{split}
\end{equation}
\textit{TotalNum} represents the number of GT or detected characters, and \textit{CorrectNum} denotes the number of correct characters. \textit{GranulPenalty} is proportional to the split number of GT or detection boxes. Each attribute will be explained in the following subsections.

\subsubsection{\textit{TotalNum}: Total Number of Characters}

\textit{TotalNum}, the denominator in Eq. ~\ref{eq:score_policy}, indicates the number of target characters. When evaluating the recall of $\text{G}_i$, $\textit{TotalNum}_i^G$ is set to the GT text length, $l^{G}_{i}$. However, note that the value of text length is absent when measuring detector accuracy. The value of  $\textit{TotalNum}_j^D$ differs depending on the availability of word transcriptions.

For end-to-end evaluation, $l^{D}_{j}$, which is the length of the predicted text in $\text{D}_j$, can be used to represent $\textit{TotalNum}_j^{D}$. However, for detection evaluation, the length of predicted word transcription is unknown. In this case, we define $\textit{TotalNum}_j^{D}$ using $\mathcal{d}_{j}$ in Table. \ref{tab:statistics}, where it defines the number of included PCC points of all the matched GT boxes.

\begin{table*}[h]
    \tabcolsep=0.16cm 

    \fontsize{8.5}{8.5}\selectfont
    \renewcommand*{\arraystretch}{1.5}
    \centering
    \begin{tabular}{c|c|c||c|c|c|c||c|c|c|c}
    \multicolumn{3}{c||}{\textbf{Evaluation}} &
    \multicolumn{4}{c||}{\textbf{Detection}} & \multicolumn{4}{c}{\textbf{End-to-end}}\\
    \hline
    \multicolumn{3}{c||}{\textbf{Score}} &
    \multicolumn{2}{c|}{\textbf{Recall}} & \multicolumn{2}{c||}{\textbf{Precision}} & \multicolumn{2}{c|}{\textbf{Recall}} & \multicolumn{2}{c}{\textbf{Precision}}\\
    \hline\hline
    \multirow{2}{*}{\rotatebox[origin=c]{90}{Granularity}} &
    Split &
    \raisebox{-0.45\totalheight}{\includegraphics[width=14mm]{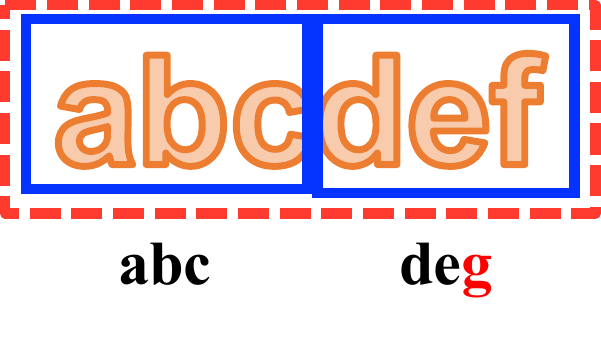}} &
    $\dfrac{6 - 1}{6}$ &
    \boldmath{$\dfrac{5}{6}$} &
    $\dfrac{3 - 0}{3} \oplus \dfrac{3 - 0}{3}$ & \boldmath{$\dfrac{6}{6}$} &
    $\dfrac{5 - 1}{6}$ &
    \boldmath{$\dfrac{4}{6}$} &
    $\dfrac{3 - 0}{3} \oplus \dfrac{2 - 0}{3}$ & \boldmath{$\dfrac{5}{6}$}\\
    \cline{2-11}
    & Merge & 
    \raisebox{-0.45\totalheight}{\includegraphics[width=14mm]{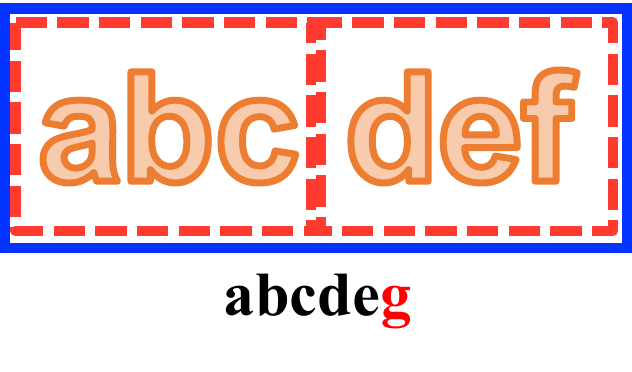}} &
    $\dfrac{3-0}{3} \oplus \dfrac{3-0}{3}$ & \boldmath{$\dfrac{6}{6}$} &
    $\dfrac{6-1}{6}$ & \boldmath{$\dfrac{5}{6}$} &
    $\dfrac{3-0}{3} \oplus \dfrac{2-0}{3}$ & \boldmath{$\dfrac{5}{6}$} &
    $\dfrac{5-1}{6}$ & \boldmath{$\dfrac{4}{6}$} \\
    
    \hline
    
    \multirow{2}{*}{\rotatebox[origin=c]{90}{Correctness}} &
    Overlapping & 
    \raisebox{-0.45\totalheight}{\includegraphics[width=14mm]{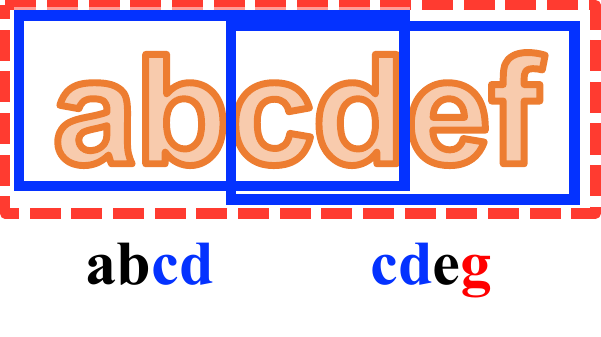}} &
    $\dfrac{6-1}{6}$ &
    \boldmath{$\dfrac{5}{6}$} &
    $\dfrac{3-0}{4} \oplus \dfrac{3-0}{4}$ & \boldmath{$\dfrac{6}{8}$} &
    $\dfrac{5-1}{6}$ &
    \boldmath{$\dfrac{4}{6}$} &
    $\dfrac{4-0}{4} \oplus \dfrac{1-0}{4}$ & \boldmath{$\dfrac{5}{8}$} \\
    \cline{2-11}
    & Missing & 
    \raisebox{-0.45\totalheight}{\includegraphics[width=14mm]{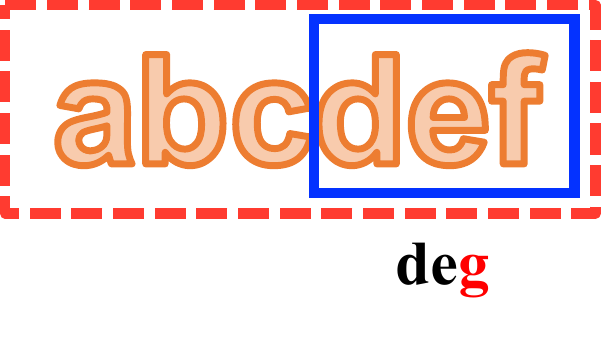}} &
    $\dfrac{3-0}{6}$ & \boldmath{$\dfrac{3}{6}$} &
    $\dfrac{3-0}{3}$ &
    \boldmath{$\dfrac{3}{3}$} &
    $\dfrac{2-0}{6}$ & \boldmath{$\dfrac{2}{6}$} &
    $\dfrac{2-0}{3}$ &
    \boldmath{$\dfrac{2}{3}$}\\
    
    \hline
    \rotatebox[origin=c]{90}{Etc} &
    FalsePositive &
    \raisebox{-0.45\totalheight}{\includegraphics[width=15mm,height=11mm]{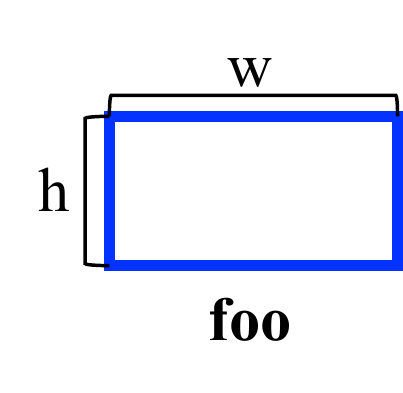}}&
    $\dfrac{0-0}{0}$ & \boldmath{$\dfrac{0}{0}$} & $\dfrac{0-0}{round(w/h)}$ & \boldmath{$\dfrac{0}{3}$} &
    $\dfrac{0-0}{0}$ & \boldmath{$\dfrac{0}{0}$} & $\dfrac{0-0}{3}$ & \boldmath{$\dfrac{0}{3}$} \\
    \hline\hline
    \multicolumn{3}{c||}{\textbf{Scoring policy}} & \multicolumn{8}{c}{ \scalebox{1.0}{$\textbf{Recall} = \frac{ \sum_{i=1}^{\left| G\right|}{( \textit{CorrectNum}_{i}^{G} - \textit{GranulPenalty}_{i}^{G}})}{ \sum_{i=1}^{\left| G\right|}{\textit{TotalNum}^G_i } }$, $\textbf{Precision} = \frac{ \sum_{j=1}^{\left| D\right|}{( \textit{CorrectNum}_{j}^{D}  - \textit{GranulPenalty}_{j}^{D}})}{ \sum_{j=1}^{\left| D\right|}{\textit{TotalNum}^D_j } }$}}\\
    \hline
    
    \end{tabular}
    \caption{Comprehensive examples of scoring for issues in text evaluations. Note that the scores are expressed as fractions rather than reduction because each denominator has the important meaning; that is character length. $\oplus$ indicates the separated summation of each nominator and denominator.}
    \vspace{-3mm}
    \label{tab:scoring_examples}
\end{table*}

\subsubsection{\textit{CorrectNum}: Correct Number of Characters}


\noindent \textbf{Correct Number for end-to-end evaluation}
Since word transcriptions are available when performing an end-to-end evaluation, the number of correct characters can be measured by finding a subsequence between the transcriptions of matched GTs and detection boxes. However, multiple matches could occur during the instance matching process, and thus, a character score could be calculated multiple times. To avoid this problem, we introduce \textit{Subsequence Elimination Scoring Process (SESP)} that calculates each character score once and eliminates the matched subsequences in both GTs and predictions.

\begin{figure}[h]
    \centering
    \includegraphics[width=0.9\linewidth]{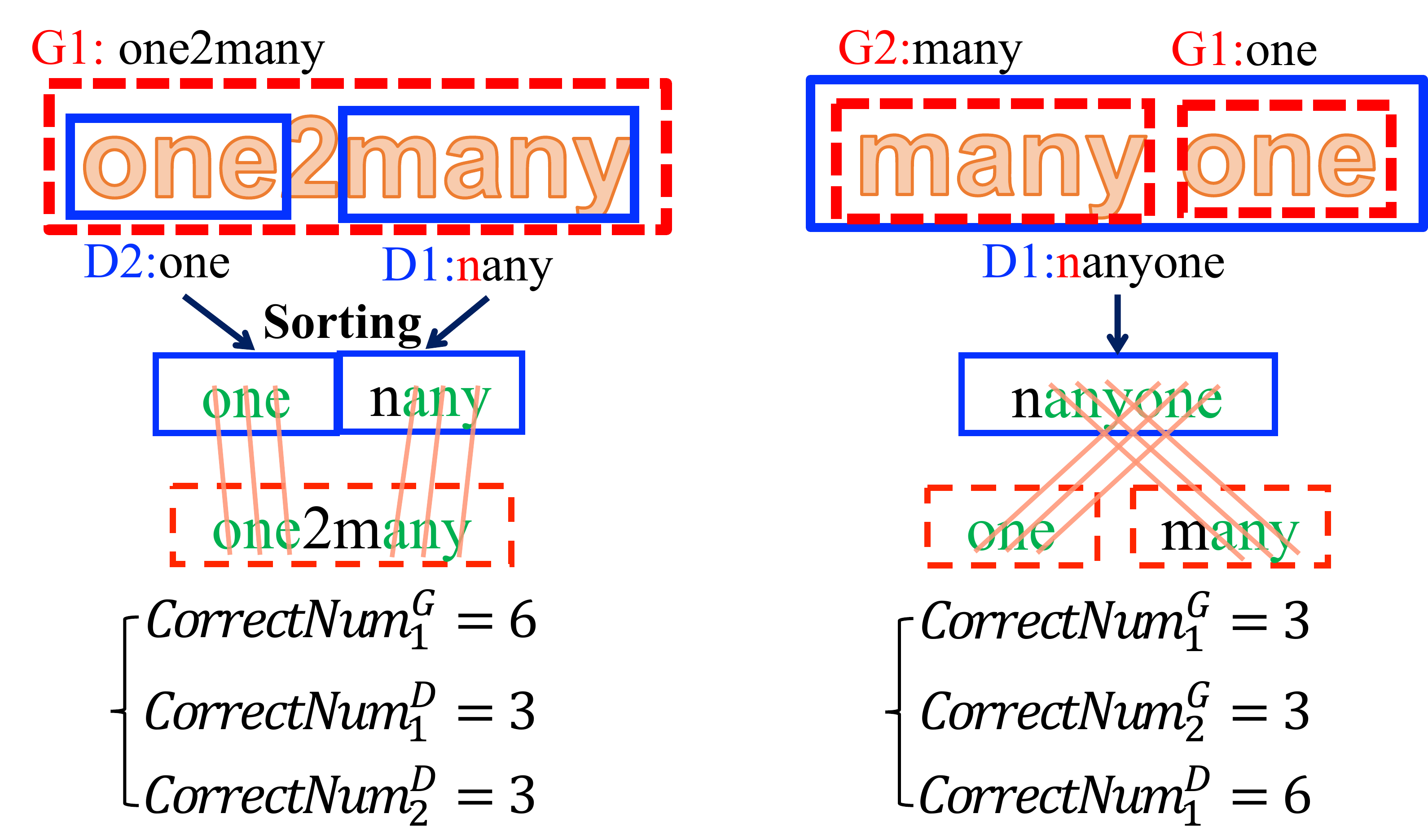}
    \caption{Examples of SESP on different matched cases: split case on left, merge case on right.}
    \label{fig:subsequence_elimination}
\end{figure}

SESP is described in Algorithm~\ref{algo:sesp}. For each GT box, a set of matched detection boxes are collected and sorted according to the order of included PCC points. Given sorted detection boxes, word transcriptions are assembled together to form a single word($\textit{recog\_text}$). We then extract $\textit{common\_seq}$, which is the Longest Common Sequence(LCS)\cite{LCS-ED-Relation} between GT and $\textit{recog\_text}$. The length of $\textit{common\_seq}$ is directly used as ${CorrectNum}_i^G$.
For each matched detection box, $\textit{det\_seq}$ is extracted between the $\textit{common\_seq}$ and $\text{D}_j$, and the length of $\textit{det\_seq}$ within all matched GTs is accumulated to ${CorrectNum}_j^D$. Finally, the  $\textit{det\_seq}$ gets eliminated in both $\text{D}^{text}_j$ and $\textit{common\_seq}$. This elimination process is required to avoid multiple matches between detection and GT transcriptions. Figure \ref{fig:subsequence_elimination} shows an example of SESP on split and merge cases.

Note that unlike the PopEval\cite{lee2019popeval}, which ignores the order of characters for scoring, we perform SESP after ordering detection boxes into the right sequence. Therefore, when performing evaluation, the order of the characters are taken into account. Our metric is almost free from having errors due to character permutations.

\begin{algorithm}[h]
    \fontsize{8}{8}\selectfont
    \SetKwFunction{hi}{hi}
    \SetKwFor{ForGT}{for}{in $\boldsymbol{G}$}{end}
    \SetKwFor{ForDS}{for}{in $\boldsymbol{DG_{i}}$}{end}

    \SetAlgoLined
    \ForGT{$\text{G}_i$}{
        $\boldsymbol{DG_{i}}$  $\leftarrow$ a set of the matched $\text{D}_j$s with $\text{G}_i$ \\
        \textit{recog\_text} $\leftarrow$ a serialized text from $\text{D}_j$ in $\boldsymbol{DG_{i}}$ \\ \hspace{0.25\linewidth} in the order of the matched PCCs \\ 
        \textit{common\_seq} $\leftarrow$ the Longest Common Subsequence(LCS) \\ \hspace{0.3\linewidth} between ($\text{G}^{\text{text}}_i$, \textit{recog\_text} ) \\
        
        ${CorrectNum}_i^G \leftarrow $ len(\textit{common\_seq}) \\
        \ForDS{$\text{D}_j$}{
            \textit{det\_seq} $\leftarrow$ the characters from $\text{D}_j$ in \textit{common\_seq} \\
            ${CorrectNum}_j^D += $ len(\textit{det\_seq}) \\
            $\text{D}^{\text{text}}_j -=$ \textit{det\_seq} \\
            \textit{common\_seq} $-=$ \textit{det\_seq} \\
        } 
        
    }
\caption{Subsequence Elimination Scoring Process (SESP)}
\label{algo:sesp}
\end{algorithm}

\noindent \textbf{Correct Number for detection evaluation}

Word transcription is not available when evaluating detection results. Therefore, we utilize the number of PCC inclusion since the detection accuracy is related to whether a detected box covers a character or not. The number of correct characters, \textit{CorrectNum}, is defined using detection statistics as follows;
\begin{equation}\label{eq:correct_length}
\begin{split}
\textit{CorrectNum}^G_{i} & = \sum_{k=1}^{l_i} \mathbb{I}\left(\mathcal{g}^{k}_{i} \geq 1 \right), \\
\textit{CorrectNum}^D_{j} & = \sum_{i}\sum_{k=1}^{l_i} \frac{\mathbb{I}\left(\mathcal{g}^{k}_{i} \geq 1 \right)}{\max(\mathcal{g}^{k}_{i}, 1)}.
\end{split}
\end{equation}
$\textit{CorrectNum}_i^G$ indicates the number of included PCC points of $\text{G}_i$ within detection boxes, and $\textit{CorrectNum}_j^D$ represents the number of accumulated PCC points of $\text{D}_j$ within all the matched $\text{G}_i$s. Additionally, $\textit{CorrectNum}j^D$ of each character is divided by the inclusion counts $\mathcal{g}^{k}_{i}$ to penalize overlapping cases. By doing this, only one of the matched characters is marked correct, and this can be seen from the same perspective of the subsequence elimination process when measuring end-to-end results.


\subsubsection{\textit{GranulPenalty}: Granularity Penalty}

The granularity indicates the connectivity condition between characters. We define \textit{GranulPenalty} as a penalty representing how much the detection result loses the connectivity information.

From a GT perspective, the most ideal condition is formed when a single detection box is matched($\mathcal{G}_{i}=1$). Likewise, from a detection perspective, the most ideal condition is formed when a single GT box is matched($\mathcal{D}_{j}=1$). The granularity penalty equation is shown in Eq. \ref{eq:grad_penalty}. As number of $\mathcal{D}_{j}$ and $\mathcal{G}_{i}$ grows, penalty increases proportionally.
\begin{equation}\label{eq:grad_penalty}
\begin{gathered}
    \textit{GranulPenalty}^G_{i}=\mathcal{G}_{i} - 1, \\
    \textit{GranulPenalty}^D_{j}=\mathcal{D}_{j} - 1.
\end{gathered}
\end{equation}
This equation means that the weight of the loss of  connectivity is same as the failure to detect a single character.


\vspace{-0.5mm}
\subsubsection{Character Number of False Positive Detection}
An appropriate penalty should be given to false positive(FP) detection, but we can't get the number of characters to penalize FP detection explicitly for detection evaluation. Therefore, the character length of the FP is estimated by assuming that the number of characters is proportional to the aspect ratio to fit into the box. As a result, the \textit{TotalNum} for FP is given through aspect ratio as shown in Eq. \ref{eq:fp_penalty_a}. For end-to-end evaluation, the character length of recognized text $l_{j}^{D}$ in the detection box is given, so this value is applied to the \textit{TotalNum} of FP as shown in Eq. \ref{eq:fp_penalty_b}.
\begin{subequations}\label{eq:fp_penalty}
\begin{equation}\label{eq:fp_penalty_a}
    \textit{TotalNum}^D_{j|\mathcal{D}_{j} = 0 } =round(w/h)
\end{equation}
\begin{equation}\label{eq:fp_penalty_b}
    \textit{TotalNum}^D_{j|\mathcal{D}_{j} = 0 } =l_{j}^{D}
\end{equation}
\end{subequations}
where $h$ indicates the minimum length of bounding box of the detection, and $w$ indicates the maximum length of it.

\subsubsection{Scoring summary}
Table ~\ref{tab:scoring_examples} shows how scoring is processed on various issues. Basically, instance matching is processed on both detection and end-to-end evaluations. The scoring process, however, differs depending on the level of evaluation. When estimating detection performance, we take the information of inclusive PCC points, and when evaluating end-to-end results, we take the correct subsequence of recognized texts.

In this way, recall and precision of each box instance are obtained. Other evaluation metrics calculate the final score by taking the average of all recall and precision values. This is not the case of our evaluation since the denominator needs to be the sum of character numbers. The final recall and precision values are obtained by separately adding numerator and denominator scores of each instance as
\begin{equation}
	\begin{gathered}
	\begin{split}
        \textit{Recall} & = \frac{ \sum_{i=1}^{\left| G\right|}{( \textit{CorrectNum}_{i}^{G} - \textit{GranulPenalty}_{i}^{G}})}{ \sum_{i=1}^{\left| G\right|}{\textit{TotalNum}^G_i } }, \\
        \textit{Precision} & = \frac{ \sum_{j=1}^{\left| D\right|}{( \textit{CorrectNum}_{j}^{D}  - \textit{GranulPenalty}_{j}^{D}})}{ \sum_{j=1}^{\left| D\right|}{\textit{TotalNum}^D_j } }.
    \end{split}
    \end{gathered}
\end{equation}
Finally, \textit{H-Mean} is calculated using Eq. \ref{eq:hmean} as usual.
\begin{equation}\label{eq:hmean}
        \textit{H-Mean} = 2\times\frac{\textit{Recall} \times \textit{Precision}}{\textit{Recall} + \textit{Precision}}
\end{equation}
Explicit recognition performance is also important for researchers developing recognition models. Using the attributes introduced in section 3.1, we measure the sole performance of the recognizer.

In order to solely evaluate recognition outputs, it is necessary to eliminate factors coming from detection outputs. One element that does not affect recognition performance is box granularity. We therefore remove granularity penalty when evaluating recognition performance. Additionally, unpaired prediction boxes should also be excluded. End-to-end performance assigns a penalty if a predicted word has no GT pair, and this is not fair since the error propagates from the detection performance. After eliminating the factors that disrupt fair recognition performance, we obtain Eq. \ref{eq:rod} that expresses the Recognition Score \textit{(RS)}.
\begin{equation}\label{eq:rod}
    \textit{RS} = \frac{ \sum_{j=1}^{\left| D\right|}{\textit{CorrectNum}_{j}^{D}} \times \mathbb{I}\left(\mathcal{D}_{j} > 0 \right)}{ \sum_{j=1}^{\left| D\right|}{ max ( \textit{TotalNum}_{j}^{D}, \mathcal{d}_{j}) }  \times \mathbb{I}\left(\mathcal{D}_{j} > 0 \right)}
\end{equation}
The equation measures recognition performance by dividing the number of correctly recognized characters by the total number of predicted characters matched with the GT instance.

\vspace{-1.0mm}
\section{Experiments}
\vspace{-1.0mm}

In this section, for ease of discussion, we analyze the tendency of our metric using the toy-examples constructed on ICDAR2013 dataset. The evaluation of our metric on real detection and recognition outputs are provided in the appendix.

\subsection{Toy-example experiments}
\begin{figure}[h]
  \centering
  \includegraphics*[width=.95\linewidth]{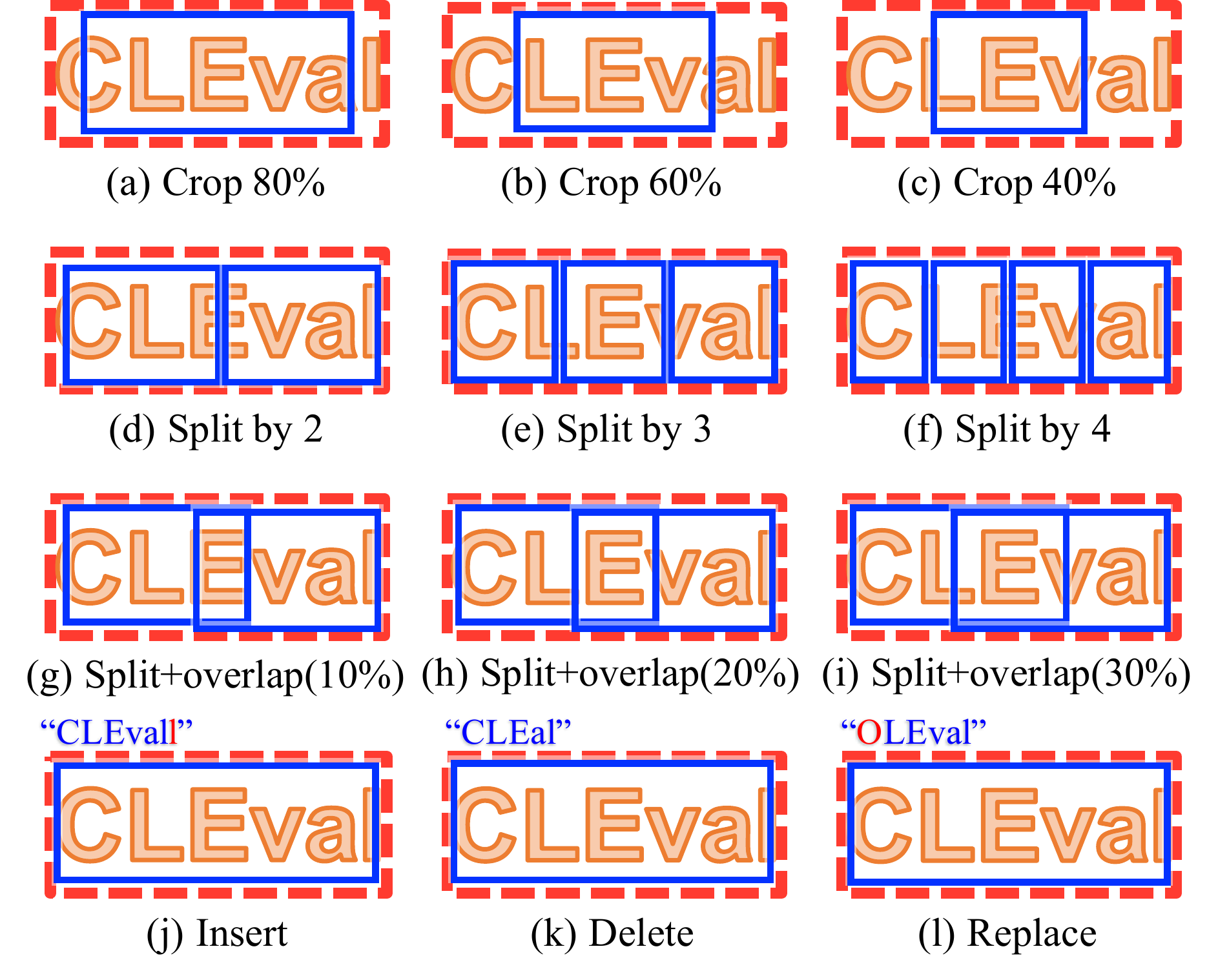}
  \caption{Detailed cases for toy-example experiment. First row shows crop cases, second row shows split cases, and third row shows overlap cases. Recognition cases such as insert, delete, and replace are in the last row. The number of insertion, deletion and replacement is increased from 1 to 3.}
  \label{fig:toy_example}
  \vspace{-3mm}
\end{figure}

To compare the characteristics of the evaluation metric, a toyset is designed to reflect the granularity and correctness issues. To evaluate detection performance, nine cases were synthetically generated using ICDAR2013 dataset\cite{karatzas2013icdar}. The cases are categorized into three parts; crop, split and overlap. To simulate recognition issues, we expect that the synthetic detection results have the same box as the GTs, and modify the text to cover insert, delete, and replace cases. Detailed toy-examples are illustrated in Figure~\ref{fig:toy_example}.

\subsection{Detection evaluation}

The detection evaluation result of the toy-example experiment is shown in Table~\ref{tab:toyset_result_det}. DetEval and IoU metrics show typical problems encountered when using a threshold based binary scoring policy. For example, DetEval uses an area recall threshold value of 0.8. Any detection boxes outside this threshold are not considered a match, and therefore, the H-mean values under crop ratio 80\% gets a value close to 0. Similar tendency is also found in the IoU metric. The metric uses a threshold value of 0.5, and thus, the H-mean value under crop ratio 50\% gets a value close to 0. This indicates that the binary scoring process does not take into account boxes that do not meet predefined threshold conditions.

As shown in Table. ~\ref{tab:toyset_result_det}, DetEval and IoU metric produces unreasonable values in many cases. DetEval scores in split and overlap cases are almost identical. We expect the scores to be different, but we get the same results because a penalty of 0.8 is assigned in all cases. On the other hand, the IoU recall precision values are strange in split and overlap cases. This is because only one of the detection boxes is paired with the GT box. Recall value of the matched detection box is close to 1 and the precision value of the mismatched detection box is close to 0.5. Also, the DetEval and IoU scores remain the same regardless of the change in overlapping ratio.

While DetEval and IoU metrics fail to cover acceptable detection results, CLEval metric performs fine-grained evaluation on detection results. The calculated recall score in CLEval is proportional to the size of the cropped box region. For the overlapping case, a precision penalty relative to the size of the overlapping region is given. The recall scores in three overlapping cases are almost the same. This is reasonable because every GT character is detected, and the number of detected duplicate characters decrease the precision score.

During the CLEval evaluation process, intermediate attributes directly related to \textit{CorrectNum}, \textit{TotalNum}, and \textit{GranulPenalty} are extracted. These are the number of split and merge, frequency of missing and overlapping characters, and estimated character numbers in false positives. Table ~\ref{tab:toyset_attribute} shows a summary of extracted intermediate attributes on ICDAR2013 toy dataset. Each quantified attribute conveys a practical view to researchers and end-users. The information can be used by the researchers to further  analyze and develop detection models.

\begin{table}[t]
    \tabcolsep=0.1cm
    \fontsize{8.5}{8.5}\selectfont
    \centering
    \begin{tabular}{c||c|c|c|c|c|c|c|c|c}
        \hline
        \rule{0pt}{9pt} \multirow{3}{*}{ \textbf{Case} } & \multicolumn{9}{c}{ \textbf{Detection Metrics} } \\
        \cline{2-10}
        \rule{0pt}{9pt} & \multicolumn{3}{c|}{DetEval} & \multicolumn{3}{c|}{IoU} & \multicolumn{3}{c}{CLEval} \\
        \cline{2-10}
        \rule{0pt}{9pt} & R & P & H & R & P & H & R & P & H\\
        \cline{2-10}
        \hline \hline
        Original & \RED{99.7} & \RED{99.9} & \RED{99.8} & \RED{99.8} & \RED{100} & \RED{99.9} & \RED{99.7} & \RED{98.7} & \RED{99.2} \\
        \cline{1-10}
        Crop 80\% & \RED{97.1} & \RED{97.3} & \RED{97.2} & \RED{99.8} & \RED{100} & \RED{99.9} & 82.7 & \RED{98.7} & 90.0 \\
        Crop 60\% & \BLUE{0.3} & \BLUE{0.5} & \BLUE{0.4} & \RED{99.7} & \RED{99.9} & \RED{99.8} & 60.7 & \RED{98.9} & 75.2 \\
        Crop 40\% & \BLUE{0.1} & \BLUE{0.1} & \BLUE{0.1} & \BLUE{0.0} & \BLUE{0.0} & \BLUE{0.0} & 40.1 & \RED{95.2} & 56.4 \\
        \cline{1-10}
        Split by 2 & 78.7 & 79.9 & 79.3 & \RED{98.4} & 49.3 & 65.7 & 82.4 & \RED{97.2} & 89.2 \\
        Split by 3 & 78.6 & 79.8 & 79.2 & \BLUE{0.0} & \BLUE{0.0} & \BLUE{0.0} & 66.7 & 94.2 & 78.1 \\
        Split by 4 & 78.5 & 79.7 & 79.1 & \BLUE{0.0} & \BLUE{0.0} & \BLUE{0.0} & 53.2 & 89.9 & 66.8 \\
        \cline{1-10}
        Overlap 10\% & 78.9 & 79.8 & 79.4 & \RED{99.8} & 50.0 & 66.6 & 81.1 & 88.7 & 84.7 \\
        Overlap 20\% & 79.0 & 79.9 & 79.4 & \RED{99.8} & 50.0 & 66.6 & 80.9 & 82.0 & 81.5 \\
        Overlap 30\% & 79.0 & 79.8 & 79.4 & \RED{99.8} & 50.0 & 66.6 & 80.9 & 74.0 & 77.3 \\
        \hline
    \end{tabular}
    \caption{Comparison of detection evaluation metrics on toy-example from ICDAR2013 dataset. Some scores are highlighted: \RED{Red} above 95, \BLUE{Blue} below 5.}
    \label{tab:toyset_result_det}
    \vspace{-5mm}
\end{table}

\begin{table}[h]
    \fontsize{8.5}{8.5}\selectfont
    \renewcommand*{\arraystretch}{1.0}
    \centering
    \begin{tabular}{c||c|c|c|c|c}
        \hline
        \rule{0pt}{9pt} \multirow{2}{*}{ \textbf{Case} } & \multicolumn{5}{c}{ \textbf{Attibutes from CLEval} } \\
        \cline{2-6}
        \rule{0pt}{9pt} & Split & Merge & Miss & Overlap & FP\\
        \cline{2-6}
        \hline \hline
        Original & 15 & 9 & 0 & 61 & 0\\
        \hline
        Crop 80\% & 15 & 9 & 996 & 50 & 0\\
        Crop 60\% & 12 & 8 & 2292 & 27 & 0\\
        Crop 40\% & 9 & 7 & 3499 & 12 & 96\\
        \hline
        Split by 2& 1014 & 12 & 0 & 60 & 93\\
        Split by 3& 1014 & 16 & 0 & 61 & 276\\
        Split by 4& 1014 & 19 & 0 & 60 & 581\\
        \hline
        Overlap 10\%& 1085 & 15 & 0 & 710 & 10\\
        Overlap 20\%& 1093 & 15 & 0 & 1252 & 2\\
        Overlap 30\%& 1093 & 15 & 0 & 2020 & 2\\
        \hline
    \end{tabular}
    \caption{Detection attributes from CLEval on ICDAR2013 dataset.}
    \label{tab:toyset_attribute}
    \vspace{-5mm}
\end{table}

\subsection{End-to-end evaluation}

In end-to-end evaluation, the strength of using CLEval metric is much more apparent. We insert, delete, and replace characters in GT transcriptions to form end-to-end test samples. When using IoU+CRW metric, all H-mean values become 0 since CRW fails to evaluate partially recognized texts. On the other hand, CLEval metric assigns partial scores according to the conditions. In the case of insertion, recall value becomes 1 because predicted transcription contains all GT characters. In the case of deletion, precision value becomes 1 because recognized texts are marked all correct. In the case of replacement, the score is affected by the penalty added to the character that is incorrectly recognized.

While performing CLEval end-to-end evaluation, we can also obtain Recognition Score (RS). The RS value is obtained regardless of the detection result by mathematically removing the detection-related terms. A detailed description of RS is provided in the appendix with actual examples.




\begin{table}[t]
    \tabcolsep=0.2cm
    \fontsize{8.5}{8.5}\selectfont
    \centering
    \begin{tabular}{c||c|c|c|c|c|c|c}
        \hline
        \rule{0pt}{9pt} \multirow{3}{*}{ \textbf{Case} } & \multicolumn{7}{c}{ \textbf{End-to-end Metrics} } \\
        \cline{2-8}
        \rule{0pt}{9pt} & \multicolumn{3}{c|}{IoU + CRW} & \multicolumn{4}{c}{CLEval} \\
        \cline{2-8}
        \rule{0pt}{9pt} & R & P & H & R & P & H & RS\\
        \cline{2-8}
        \hline \hline
        Original & \RED{99.6} & \RED{99.8} & \RED{99.7} & \RED{99.7} & \RED{99.7} & \RED{99.7} & \RED{98.9} \\
        \hline
        Insert 1 & \BLUE{0.0} & \BLUE{0.0} & \BLUE{0.0} & \RED{99.7} & 84.0 & 91.2 & 83.6 \\
        Insert 2 & \BLUE{0.0} & \BLUE{0.0} & \BLUE{0.0} & \RED{99.7} & 73.1 & 84.4 & 73.1 \\
        Insert 3 & \BLUE{0.0} & \BLUE{0.0} & \BLUE{0.0} & \RED{99.7} & 65.7 & 79.2 & 65.6 \\
        \hline
        Delete 1 & \BLUE{0.0} & \BLUE{0.0} & \BLUE{0.0} & 81.0 & \RED{99.7} & 89.4 & 80.4 \\
        Delete 2 & \BLUE{0.0} & \BLUE{0.0} & \BLUE{0.0} & 63.7 & \RED{99.6} & 77.7 & 63.3 \\
        Delete 3 & \BLUE{0.0} & \BLUE{0.0} & \BLUE{0.0} & 48.0 & \RED{99.5} & 64.8 & 47.8 \\
        \hline
        Replace 1 & \BLUE{0.0} & \BLUE{0.0} & \BLUE{0.0} & 81.0 & 81.0 & 81.0 & 80.4 \\
        Replace 2 & \BLUE{0.0} & \BLUE{0.0} & \BLUE{0.0} & 64.1 & 64.2 & 64.1 & 63.7 \\
        Replace 3 & \BLUE{0.0} & \BLUE{0.0} & \BLUE{0.0} & 49.9 & 49.9 & 49.9 & 49.7 \\
        \hline
    \end{tabular}
    \caption{Comparison of end-to-end metrics on toy-example. Some scores are highlighted: \RED{Red} above 95, \BLUE{Blue} below 5.}
    \label{tab:toyset_result_e2e}
    \vspace{-5mm}
\end{table}

\section{Conclusion}
Fair and detailed evaluation of OCR models is needed, and yet, no robust evaluation metric was proposed in the OCR community. The proposed CLEval metric could evaluate text detection, recognition, and end-to-end results. This is done by solving the granularity and correctness issues by performing instance matching and character scoring process. Our metric allows fine assessment, and alleviates qualitative disagreement. We expect researchers and end-users to take advantage of the metric to conduct thorough end-to-end evaluation.





{\small
\bibliographystyle{ieee_fullname}
\bibliography{mybib}
}

\clearpage
\appendix
\section{Toy-example experiment on ICDAR2015.}

\begin{figure*}[ht!]
 \centering
 \hfill\includegraphics*[width=.6\linewidth]{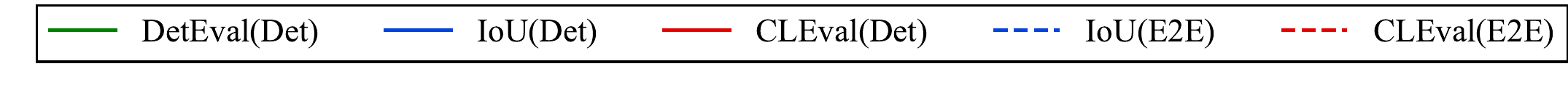}
 \begin{subfigure}{.31\linewidth}
 \centering
 \includegraphics*[width=0.95\linewidth, clip=true]{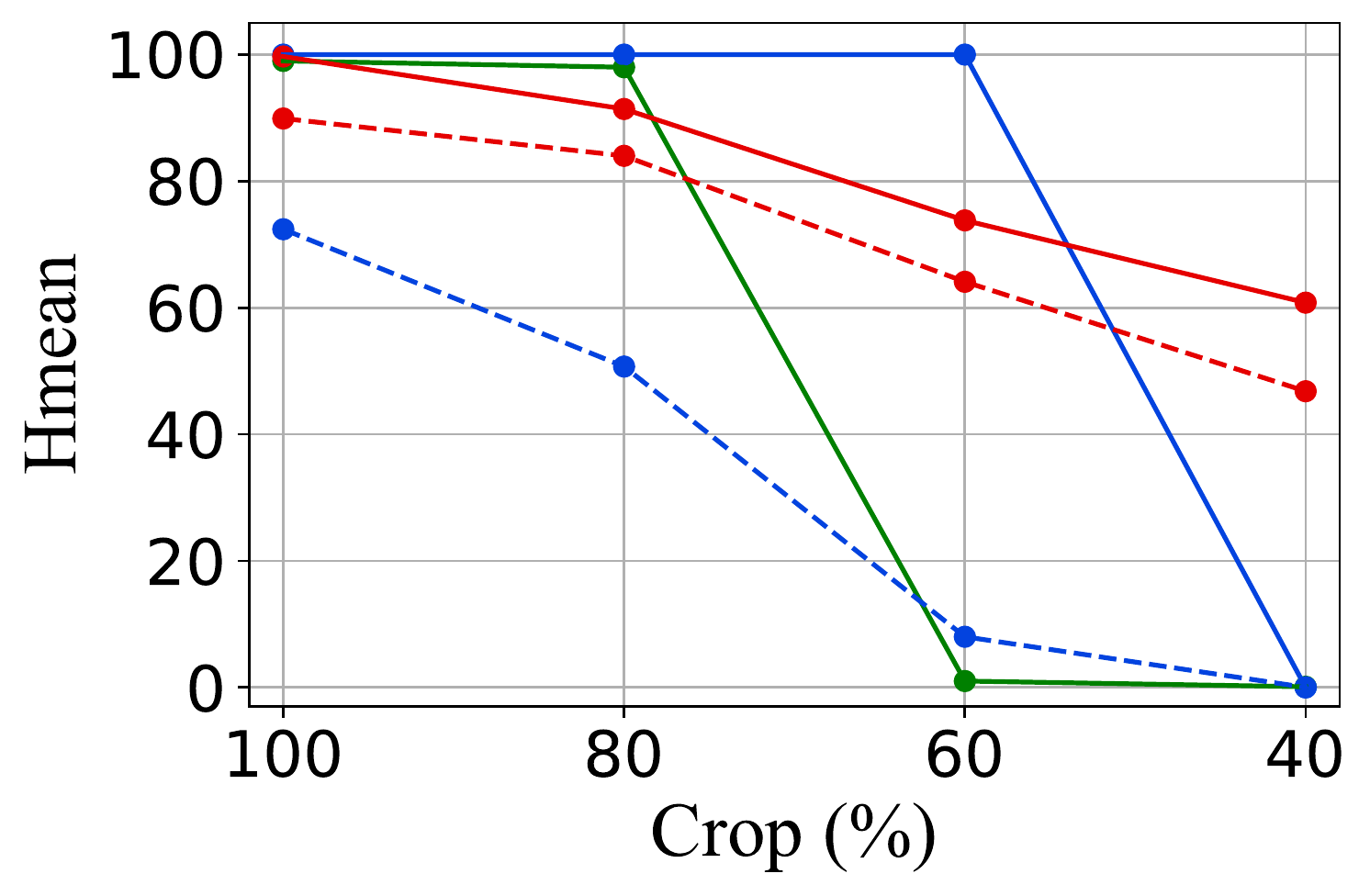}
 \end{subfigure}%
 \begin{subfigure}{.31\linewidth}
 \centering
 \includegraphics*[width=0.95\linewidth, clip=true]{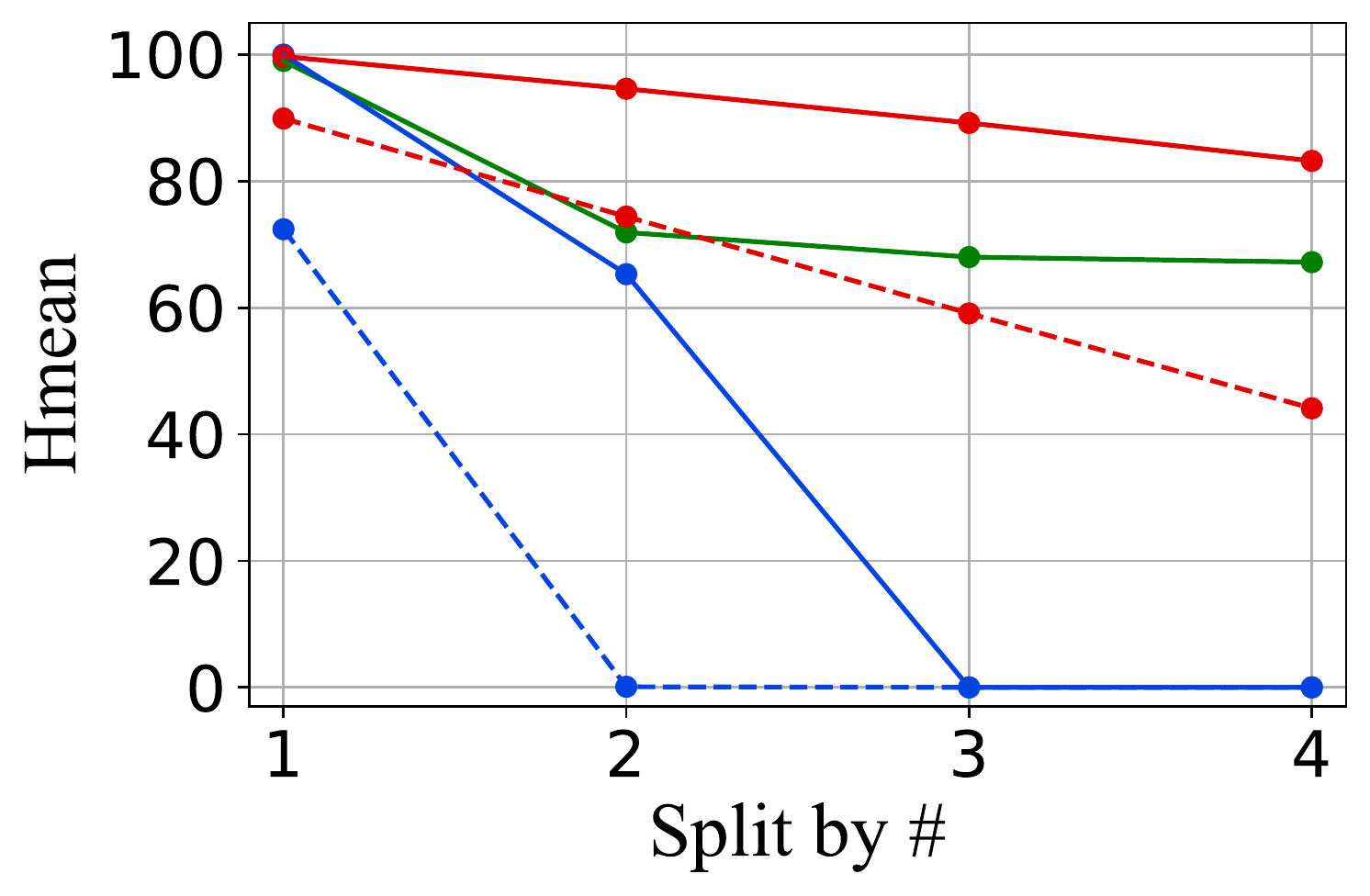}
 \end{subfigure}%
 \begin{subfigure}{.31\linewidth}
 \centering
 \includegraphics*[width=0.95\linewidth, clip=true]{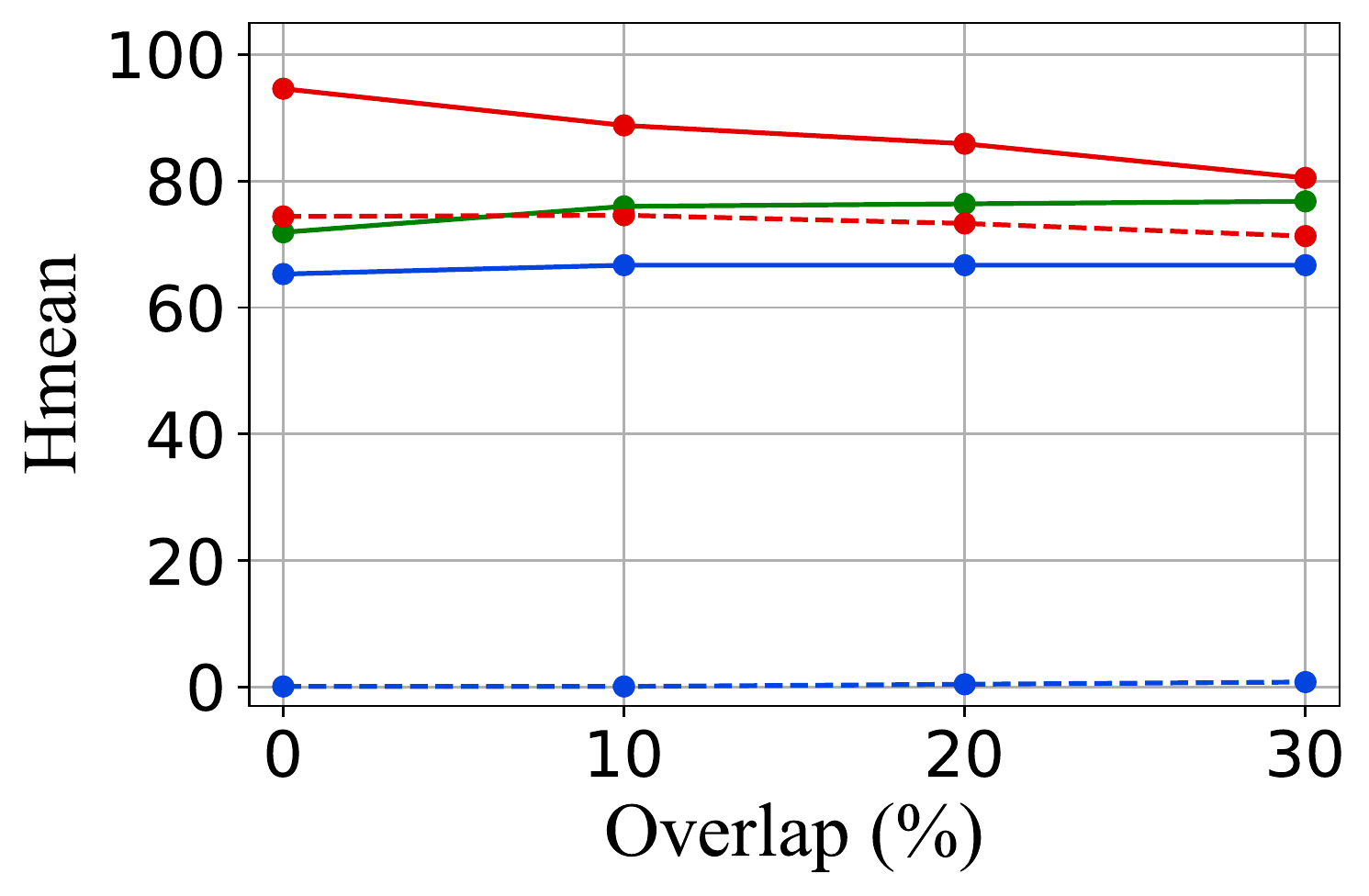}
 \end{subfigure}%
 \vspace{-3mm}
 \caption{Line graph for H-mean of each evaluation metric according to different crop, split, overlap condition. Solid line indicates the detection evaluation while dashed line indicates the end-to-end evaluation results. }
  \label{fig:toyset_hmean_graph_IC15}
  \vspace{-3mm}
\end{figure*}

To show the stability of our metric, we additionally performed experiments on ICDAR2015 dataset. For detection evaluation, toy-set was produced in the same way as it was made using the ICDAR2013 dataset, and for end-to-end evaluation we constructed another set based on detection toy-examples. Note that we evaluated the end-to-end result using the best model reported in \cite{baek2019wrong}. The result of the experiment and their attributes from CLEval are shown in table~\ref{tab:ic15_attribute},\ref{tab:toyset_result_IC15}, and the line graphs in Figure~\ref{fig:toyset_hmean_graph_IC15} show results performed under different conditions.


The results on ICDAR2015 dataset show similar tendency when compared with the evalution results on ICDAR2013 dataset. Due to the trait of the IoU metric using a threshold value of 0.5, the metric assigns zero score to the cropped area less than 50 percent. Also, the zero scores on split cases are caused by the absence of handling granularity issues. One-to-many or many-to-one match cases frequently occur, but the IoU metric only considers one-to-one matching cases. Multiple box predictions could cover a single ground truth box, but zero scores are given if the overlapping region does not meet a predefined threshold.

The same holds for the IoU+CRW metric on end-to-end evaluation. Using a predefined threshold, a one-to-one match is first made to filter out valid box candidates, then CRW is performed to identify matching transcripts. In the transcript matching process, CRW requires ground truth and predicted text to be matched perfectly. Otherwise, a zero score is assigned to the matched box candidates. For this reason, we observe meaningful comparison was difficult with the IoU+CRW.

The proposed metric provides stable scores under various cases by performing evaluations at the character-level. Table ~\ref{tab:toyset_result_IC15} shows recall, precision scores of partially corrected detections.

\begin{table}[h]
    \fontsize{9}{9}\selectfont
    \centering
    \begin{tabular}{c||c|c|c|c|c}
        \hline
        \rule{0pt}{9pt} \multirow{3}{*}{ \textbf{Case} } & \multicolumn{5}{c}{ \textbf{Attributes from CLEval} } \\
        \cline{2-6}
        \rule{0pt}{9pt} & \multicolumn{2}{c|}{instance-level} & \multicolumn{3}{c}{character-level} \\
        \cline{2-6}
        \rule{0pt}{9pt} & Split & Merge & Miss & Overlap & FP\\
        \cline{2-6}
        \hline \hline
        Original & 18 & 15 & 5 & 45 & 0\\
        \hline
        Crop 80\% & 11 & 10 & 1723 & 26 & 0\\
        Crop 60\% & 9 & 9 & 4592 & 17 & 0\\
        Crop 40\% & 8 & 8 & 6246 & 10 & 0\\
        \hline
        Split by 2& 1990 & 18 & 0 & 38 & 88\\
        Split by 3& 1988 & 19 & 0 & 36 & 178\\
        Split by 4& 1994 & 21 & 0 & 35 & 633\\
        \hline
        Overlap 10\%& 2073 & 21 & 0 & 1574 & 1\\
        Overlap 20\%& 2074 & 23 & 0 & 2431 & 0\\
        Overlap 30\%& 2074 & 23 & 0 & 4157 & 0\\
        \hline
    \end{tabular}
    \caption{Detection attributes from CLEval on ICDAR2015 dataset.}
    \label{tab:ic15_attribute}
    \vspace{-5mm}
\end{table}

\begin{table*}[ht!]
    \tabcolsep=0.15cm
    \fontsize{8}{8}\selectfont
    \renewcommand*{\arraystretch}{1.1}
    \centering
    \begin{tabular}{c||c|c|c||c|c|c||c|c|c||c|c|c||c|c|c}
        \hline
        \rule{0pt}{9pt} \multirow{3}{*}{ \textbf{Case} } & \multicolumn{9}{c||}{ \textbf{Detection Metrics} }  & \multicolumn{6}{c}{ \textbf{E2E Metrics} }  \\
        \cline{2-16}
        \rule{0pt}{9pt} & \multicolumn{3}{c||}{\textbf{DetEval}$^*$} & \multicolumn{3}{c||}{\textbf{IoU}} & \multicolumn{3}{c||}{\textbf{CLEval}} & \multicolumn{3}{c||}{\textbf{IoU+CRW}} & \multicolumn{3}{c}{\textbf{CLEval}} \\
        \cline{2-16}
        \rule{0pt}{9pt} & R & P & H & R & P & H & R & P & H & R & P & H & R & P & H \\
        \hline \hline
        Original & \RED{98.1} & \RED{99.9} & \RED{99.0} & \RED{100} & \RED{100} & \RED{100} & \RED{99.8} & \RED{99.4} & \RED{99.6} & 72.4 & 72.4 & 72.4 & 88.6 & 91.1 & 89.8 \\
        \cline{1-16}
        Crop 80\% & \RED{98.0} & \RED{98.1} & \RED{98.0} & \RED{100} & \RED{100} & \RED{100} & 84.4 & \RED{99.6} & 91.4 & 50.7 & 50.7 & 50.7 & 80.6 & 87.7 & 84.0 \\
        Crop 60\% & \BLUE{0.7} & \BLUE{1.8} & \BLUE{1.0} & \RED{100} & \RED{100} & \RED{100} & 58.6 & \RED{99.6} & 73.8 & 8.0 & 8.0 & 8.0 & 54.8 & 77.0 & 64.0 \\
        Crop 40\% & \BLUE{0.0} & \BLUE{0.1} & \BLUE{0.1} & \BLUE{0.0} & \BLUE{0.0} & \BLUE{0.0} & 43.7 & \RED{99.6} & 60.8 & \BLUE{0.0} & \BLUE{0.0} & \BLUE{0.0} & 35.2 & 69.4 & 46.7 \\
        \cline{1-16}
        Split by 2 & 69.8 & 74.1 & 71.9 & \RED{97.9} & 49.0 & 65.3 & 81.9 & \RED{98.7} & 89.5 & \BLUE{0.1} & \BLUE{0.1} & \BLUE{0.1} & 63.4 & 76.6 & 69.4 \\
        Split by 3 & 65.9 & 70.3 & 68.0 & \BLUE{0.0} & \BLUE{0.0} & \BLUE{0.0} & 64.0 & \RED{97.9} & 77.4 & \BLUE{0.0} & \BLUE{0.0} & \BLUE{0.0} & 38.4 & 62.3 & 47.5 \\
        Split by 4 & 65.2 & 69.3 & 67.2 & \BLUE{0.0} & \BLUE{0.0} & \BLUE{0.0} & 49.2 & 94.1 & 64.6 & \BLUE{0.0} & \BLUE{0.0} & \BLUE{0.0} & 15.2 & 48.1 & 23.1 \\
        \cline{1-16}
        Overlap 10\% & 73.9 & 78.2 & 76.0 & \RED{100} & 50.0 & 66.7 & 81.1 & 87.4 & 84.1 & \BLUE{0.1} & \BLUE{0.1} & \BLUE{0.1} & 66.3 & 73.6 & 69.8 \\
        Overlap 20\% & 74.3 & 78.6 & 76.4 & \RED{100} & 50.0 & 66.7 & 81.1 & 81.9 & 81.5 & \BLUE{0.7} & \BLUE{0.3} & \BLUE{0.4} & 68.3 & 69.4 & 68.8 \\
        Overlap 30\% & 74.8 & 78.8 & 76.8 & \RED{100} & 50.0 & 66.7 & 81.1 & 72.6 & 76.6 & \BLUE{1.1} & \BLUE{0.5} & \BLUE{0.7} & 69.4.7 & 65.2 & 67.2 \\
        \hline
    \end{tabular}
    \caption{Comparison of evaluation metrics on toy-set from ICDAR2015 dataset. Some scores are highlighted: \RED{Red} above 95, \BLUE{Blue} below 5. $^*$denotes our implemented code since official evaluation does not exist.}
    \label{tab:toyset_result_IC15}
    \vspace{-3mm}
\end{table*}

\begin{table*}[ht!]
    \fontsize{9}{9}\selectfont
    \renewcommand*{\arraystretch}{1.1}
    \centering
    \begin{tabular}{c||c|c|c|c|c|c||c|c|c|c|c}
    \hline
    \rule{0pt}{9pt} \multirow{3}{*}{ \textbf{Detector} } & \multicolumn{6}{c||}{ \textbf{Metrics} } & \multicolumn{5}{c}{ \textbf{Attributes} } \\
    \cline{2-12}
    \rule{0pt}{9pt} & \multicolumn{3}{c|}{ \textbf{IoU} } & \multicolumn{3}{c||}{ \textbf{CLEval} } &
    \multicolumn{2}{c|}{ \textbf{instance-level} } & \multicolumn{3}{c}{ \textbf{character-level} } \\
    \cline{2-12}
    \rule{0pt}{9pt} & \textbf{R} & \textbf{P} & \textbf{H} & \textbf{R} & \textbf{P} & \textbf{H} & \textbf{Split} & \textbf{Merge} & \textbf{Miss} & \textbf{Overlap} & \textbf{FP} \\
    \hline \hline
    CTPN~\cite{tian2016ctpn} & 83.0 & 93.0 & 87.7 & 82.5 & 84.1 & 83.3 & 14 & 231 & 1015 & 56 & 411 \\
    SegLink~\cite{shi2017seglink} & 60.0 & 73.9 & 66.2 & 74.0 & 95.4 & 83.3 & 93 & 28 & 1293 & 46 & 116 \\
    EAST~\cite{zhou2017east} & 70.7 & 81.6 & 75.8 & 84.7 & 94.2 & 89.2 & 52 & 47 & 827 & 51 & 200 \\
    RRPN~\cite{ma2017rrpn} & 87.3 & 95.2 & 91.1 & 90.2 & 95.3 & 92.7 & 43 & 35 & 521 & 75 & 140 \\
    TextBoxes++~\cite{liao2018textboxes++} & 85.6 & 91.9 & 88.6 & 92.7 & 94.1 & 93.4 & 26 & 29 & 374 & 41 & 246 \\
    FOTS~\cite{liu2018fots} & 90.4 & 95.4 & 92.8 & 94.0 & 96.4 & 95.2 & 57 & 34 & 289 & 99 & 62 \\
    MaskTextSpotter~\cite{lyu2018mask} & 88.6 & 95.0 & 91.7 & 93.9 & 97.7 & 95.7 & 26 & 23 & 325 & 24 & 69 \\
    CRAFT~\cite{baek2019craft} & 93.1 & 97.4 & 95.2 & 96.3 & 96.6 & 96.4 & 34 & 37 & 177 & 84 & 60 \\
    PMTD~\cite{liu2019pmtd} & 92.2 & 95.1 & 93.6 & 96.1 & 97.6 & 96.8 & 24 & 28 & 199 & 28 & 76 \\
    \hline
    \end{tabular}
    \caption{Comparison of evaluation metrics \& additional detection attributes for different detectors on \textbf{ICDAR2013} dataset. R, P, and H refer to recall, precision, and H-mean. Detectors are sorted from the highest score on DetEval metric.}
    \label{tab:result_detector_ic13}
    \vspace{-3mm}
\end{table*}

\begin{table*}[ht!]
    \fontsize{9}{9}\selectfont
    \renewcommand*{\arraystretch}{1.1}
    \centering
    \begin{tabular}{c||c|c|c|c|c|c||c|c|c|c|c}
        \hline
        \rule{0pt}{9pt} \multirow{3}{*}{ \textbf{Detector} } & \multicolumn{6}{c||}{ \textbf{Metrics} } & \multicolumn{5}{c}{ \textbf{Attributes} } \\
        \cline{2-12}
        \rule{0pt}{9pt} & \multicolumn{3}{c|}{ \textbf{IoU} } & \multicolumn{3}{c||}{ \textbf{CLEval} } &
        \multicolumn{2}{c|}{ \textbf{instance-level} } & \multicolumn{3}{c}{ \textbf{character-level} } \\
        \cline{2-12}
        \rule{0pt}{9pt} & \textbf{R} & \textbf{P} & \textbf{H} & \textbf{R} & \textbf{P} & \textbf{H} & \textbf{Split} & \textbf{Merge} & \textbf{Miss} & \textbf{Overlap} & \textbf{FP} \\
        \hline \hline
        CTPN~\cite{tian2016ctpn} & 51.6 & 74.2 & 60.9 & 63.2 & 93.6 & 75.4 & 75 & 103 & 3842 & 31 & 302 \\
        SegLink~\cite{shi2017seglink} & 72.9 & 80.2 & 76.4 & 79.4 & 95.1 & 86.5 & 130 & 117 & 2123 & 107 & 224 \\
        RRPN~\cite{ma2017rrpn} & 77.1 & 83.5 & 80.2 & 81.8 & 94.6 & 87.7 & 60 & 77 & 1938 & 49 & 377 \\
        EAST~\cite{zhou2017east} & 77.2 & 84.6 & 80.8 & 84.8 & 93.9 & 89.1  & 66 & 124 & 1607 & 65 & 401 \\
        TextBoxes++~\cite{liao2018textboxes++} & 80.8 & 89.1 & 84.8 & 84.2 & 95.0 & 89.3 & 35 & 52 & 1713 & 33 & 397 \\
        MaskTextSpotter~\cite{lyu2018mask} & 79.5 & 89.0 & 84.0 & 83.7 & 96.2 & 89.5 & 52 & 63 & 1751 & 62 & 224 \\
        FOTS~\cite{liu2018fots} & 87.9 & 91.9 & 89.8 & 90.0 & 97.1 & 93.4 & 74 & 69 & 1033 & 67 & 160 \\
        CRAFT~\cite{baek2019craft} & 84.3 & 89.8 & 86.9 & 90.0 & 97.4 & 93.5 & 33 & 73 & 1076 & 19 & 171 \\
        PMTD~\cite{liu2019pmtd} & 87.4 & 91.3 & 89.3 & 90.8 & 97.0 & 93.8 & 38 & 47 & 982 & 33 & 232 \\
        \hline
    \end{tabular}
    \caption{Comparison of evaluation metrics \& additional detection attributes for different detectors on \textbf{ICDAR2015} dataset. R, P, and H refer to recall, precision, and H-mean. Detectors are sorted from the highest score on IoU metric. }
    \label{tab:result_detector_ic15}
    \vspace{-3mm}
\end{table*}

\section{Evaluation of text detectors}

In this section, we compare CLEval with other commonly used evaluation metrics using the state-of-the-art text detectors. We requested the authors of various scene text detectors to provide their test results on public datasets and organized the results in Table~\ref{tab:result_detector_ic13}, \ref{tab:result_detector_ic15} for ICDAR2013 and ICDAR2015, respectively.

The strength of using the CLEval metric is in its use of additional instance-level and character-level information to calculate recall, precision, and hmean values. As shown in Table~\ref{tab:result_detector_ic13}, \ref{tab:result_detector_ic15}, even without the knowledge of recall, precision, and hmean values, we could examine the quality of the detection models by observing the attributes produced by the CLEval metric.

\section{Evaluation on real end-to-end results}
In this experiment, we take a close look into the end-to-end performance of various detector and recognizer combinations. We used the well-known detectors such as CRAFT\cite{baek2019craft}, EAST\cite{zhou2017east}, RRPN\cite{ma2017rrpn}, PixelLink\cite{deng2018pixellink}, and TextBoxes++\cite{liao2018textboxes++}. We recognized the texts of those detectors with three types of recognizers provided in \cite{baek2019wrong}. CLEval results are listed in the Table \ref{tab:e2e_evaluation}. \textit{High} indicates recognizer with TPS+ResNet+BiLSTM+Attn moduels, \textit{Mid} indicates recognizer with None+VGG+BiLSTM+CTC modules, and \textit{Low} indicates recognizer with None+VGG+None+CTC modules. We observe that RS scores in each \textit{High}, \textit{Mid}, and \textit{Low} recognition combination are similar. This infers that RS can be used to evaluate recognition performance regardless of the detection module.


\begin{table*}[t!]
    \tabcolsep=0.30cm
    \fontsize{9}{9}\selectfont
    \renewcommand*{\arraystretch}{1.2}
    \centering
    \begin{tabular}{c|c|c|c||c|c|c|c|c}
        \hline
        \rule{0pt}{9pt} \multirow{2}{*}{ \textbf{Detector}} & \multicolumn{3}{c||}{ \textbf{CLEval Det} } & \multirow{2}{*}{ \textbf{Recognizer}} &   \multicolumn{3}{c|}{ \textbf{CLEval E2E} }  & {\textbf{E2E Rec} }  \\
        \cline{2-4}\cline{6-9}
        \rule{0pt}{9pt} & R & P & H & &  R & P & H & RS \\
        \hline
        \multirow{3}{*}{RRPN\cite{ma2017rrpn}} & \multirow{3}{*}{81.8} & \multirow{3}{*}{94.6}  & \multirow{3}{*}{87.7}
        & High & 76.7 & 84.3 & 79.9 & 89.0 \\
        \cline{5-9}
        & & & & Mid & 74.0 & 82.9 & 78.2 & 86.4 \\
        \cline{5-9}
        & & & & Low & 70.4 & 82.4 & 75.9 & 83.0 \\
        \hline
        \multirow{3}{*}{EAST\cite{zhou2017east}} & \multirow{3}{*}{84.8} & \multirow{3}{*}{93.9}  & \multirow{3}{*}{89.1}
        & High & 78.4 &	83.7 & 81.0 &	88.4 \\
        \cline{5-9}
        & & & & Mid & 75.2 & 83.6 & 79.2 & 85.5 \\
        \cline{5-9}
        & & & & Low & 72.0 & 82.5 & 76.9 & 82.2 \\
        \hline
        \multirow{3}{*}{TextBoxes++\cite{liao2018textboxes++}} & \multirow{3}{*}{84.2} & \multirow{3}{*}{95.0}  & \multirow{3}{*}{89.3}
        & High & 78.1 &	86.4 & 82.0 & 90.0 \\
        \cline{5-9}
        & & & & Mid & 72.2 & 84.0 &	77.6 & 84.0 \\
        \cline{5-9}
        & & & & Low & 67.8 & 82.8 & 74.6 & 79.0 \\
        \hline
        \multirow{3}{*}{PixelLink\cite{deng2018pixellink}} & \multirow{3}{*}{89.0} & \multirow{3}{*}{96.7}  & \multirow{3}{*}{92.7}
        & High & 80.8 &	88.4 & 84.5 & 87.8 \\
        \cline{5-9}
        & & & & Mid & 78.1 & 87.3 & 82.5 & 85.3 \\
        \cline{5-9}
        & & & & Low & 73.8 & 86.1 & 79.4 & 81.0 \\
        \hline
        \multirow{3}{*}{CRAFT\cite{baek2019craft}} & \multirow{3}{*}{89.9} & \multirow{3}{*}{97.3}  & \multirow{3}{*}{93.5}
        & High & 81.6 &	88.9 &	85.1 &	88.2\\
        \cline{5-9}
        & & & & Mid & 78.4 & 87.3 & 82.6 & 85.1 \\
        \cline{5-9}
        & & & & Low & 74.4 & 86.3 & 79.9 & 81.1 \\
        \hline
    \end{tabular}
    \vspace{-3mm}
    \caption{End-to-end evaluation using CLEval for state-of-the art text detectors and recognizers.}
    \label{tab:e2e_evaluation}
    \vspace{-3mm}
\end{table*}

\section{PCC generation in polygon annotation}
Most of the text bounding boxes in public datasets are represented using four quadrilateral points. However, there exist polygon-type datasets that use multiple vertexes to tightly bound the text regions. For polygon datasets, we could acquire the center information by splitting the polygon into a sub-groups of quadrilaterals. Algorithm~\ref{algo:pcc_polygon} describes the detailed procedure for generating PCCs in polygon-type dataset. By extending PCC generation to polygon datasets, CLEval can be used to evaluate on a variety of datasets represented by both rectangles and polygons.

\begin{algorithm}[h]
    \fontsize{8}{8}\selectfont
    \definecolor{mygray}{gray}{0.6}
    \SetKwProg{Def}{def}{}{end}
    \SetKwFunction{hi}{hi}
    \SetKwFor{ForSection}{for}{from $1$ to $len({P_{top})-1}$}{end}
    \SetKwFor{ForInterpolate}{for}{from $1$ to $L_{trans}-1$}{end}
    \SetKwFor{ForPCCgen}{for}{from $1$ to $L_{trans}$}{end}
    \SetAlgoLined
    ${\{\boldsymbol{p_1}, \boldsymbol{p_2}, \dots, \boldsymbol{p_{2n}}\} \leftarrow }$ a set of even-number annotation points from left-top, clockwise order \\
    ${\boldsymbol{LEN_{transcription}} \leftarrow }$ length of transcription \\
    \Def{PCC\_{polygon}(Points, Length):}{
        initialize \textit{PCCs} as array \\
        ${PTS_{top} \leftarrow }$ a set of points above center ($=\{{Points_1}, \dots, {Points_n}\}$) \\
        ${PTS_{bottom} \leftarrow }$ a set of points below center ($=\{{Points_{n+1}}, \dots, {Points_{2n}}\}$) \\
        $CharSize$ = $len(PTS_{top})-1$ \\
        \textcolor{mygray}{
            \textit{\# to make order from left to right, reverse order of element}\\
        }
        ${reverseOrder(PTS_{bottom})}$ \\
        ${NEW_{top}, NEW_{bottom}}$ = Interpolate($PTS_{top}$, $PTS_{bottom}$, $Length$) \\ 
        
        \ForPCCgen{k}{
            $char_{tl}$ = $New^{CharSize\times(k-1)+1}_{top}$ \\ 
            $char_{tr}$ = $New^{CharSize\times(k)+1}_{top}$ \\ 
            $char_{bl}$ = $New^{CharSize\times(k-1)+1}_{bottom}$ \\ 
            $char_{br}$ = $New^{CharSize\times(k)+1}_{bottom}$ \\ 
            PCCs.append($mean({char_{tl}, char_{tr}, char_{bl}, char_{br}})$) \\
        }
        
        \textbf{return} PCCs \\
    }
    \Def{Interpolate($P_{top}$, $P_{bottom}$, $L_{trans}$):}{
        initialize \textit{$New_{top}$}, \textit{$New_{bottom}$} as array \\
        \ForSection{i}{
        $p^{tl}$, $p^{tr}$ = $i^{th}$, ${(i+1)}^{th}$ point of $P_{top}$ \\
        $p^{bl}$, $p^{br}$ = $i^{th}$, ${(i+1)}^{th}$ point of $P_{bottom}$ \\
        $New_{top}$.append($p^{tl}$) \\
        $New_{bottom}$.append($p^{bl}$) \\
            \ForInterpolate{k}{
                $n_{top}$ = $\left(1-\frac{k}{L_{trans}}\right)p^{tl} + \left(\frac{k}{L_{trans}}\right)p^{tr}$ \\
                $n_{bottom}$ = $\left(1-\frac{k}{L_{trans}}\right)p^{bl} + \left(\frac{k}{L_{trans}}\right)p^{br}$ \\
                $New_{top}$.append($n_{top}$) \\
                $New_{bottom}$.append($n_{bottom}$) \\
            }
        }
        $New_{top}$.append(last element of $P_{top}$) \\
        $New_{bottom}$.append(last element of $P_{bottom}$) \\
        \textbf{return} \textit{$New_{top}$, $New_{bottom}$} \\
    }
\caption{PCC generation process from polygon annotation}
\label{algo:pcc_polygon}
\end{algorithm}

\end{document}